\documentclass[letterpaper,dvipsnames,table]{article} 
\usepackage{aaai2026}  
\usepackage{times}  
\usepackage{helvet}  
\usepackage{courier}  
\usepackage[hyphens]{url}  
\usepackage{graphicx} 
\usepackage{natbib}  
\usepackage{caption} 
\usepackage{algorithm}
\usepackage{algorithmic}
\usepackage{booktabs}
\usepackage{newfloat}
\usepackage{listings}
\DeclareCaptionStyle{ruled}{labelfont=normalfont,labelsep=colon,strut=off} 
\floatstyle{ruled}
\newfloat{listing}{tb}{lst}{}
\floatname{listing}{Listing}
\newcommand{\our}{{USPR}}
\usepackage{subcaption} 
\usepackage{amsmath}
\usepackage{bm}
\usepackage{amssymb}
\usepackage[capitalise,nameinlink]{cleveref}
\usepackage{multirow}
\usepackage{array}

\lstdefinestyle{promptstyle_main}{
    basicstyle=\scriptsize\ttfamily,
    keywordstyle=\bfseries,          
    commentstyle=\itshape,          
    frame=tb,                       
    framesep=4pt,                   
    breakautoindent=false,
    breakindent=0pt, 
    breakatwhitespace=true,
    breaklines=true,
    captionpos=b,
    columns=flexible,
    extendedchars=true,
    fontadjust=true,
    inputencoding=utf8,
    keepspaces=true,
    lineskip=-0.1pt,
    linewidth=0.98\columnwidth,
    resetmargins=true,
    showspaces=false,
    showstringspaces=false,
    showtabs=false,
    tabsize=2,
    upquote=true,
    moredelim=[s]{\{}{\}},
    morekeywords={Index, Count}, 
    morecomment=[l]{<},         
    numbers=none,
    xleftmargin=8pt,
    xrightmargin=8pt,
    aboveskip=0.8\baselineskip,
    belowskip=0.8\baselineskip
}

\lstdefinestyle{promptstyle_appendix}{
    backgroundcolor=\color{gray!8},
    basicstyle=\scriptsize\ttfamily\color{black!90},
    breakautoindent=false,
    breakindent=0pt, 
    breakatwhitespace=true,
    breaklines=true,
    captionpos=b,
    keepspaces=true,
    showspaces=false,
    showstringspaces=false,
    showtabs=false,
    frame=single,
    rulecolor=\color{gray!40},
    framesep=3pt,
    frameround=tttt,
    framexleftmargin=6pt,
    xleftmargin=8pt,
    xrightmargin=8pt,
    tabsize=2,
    linewidth=0.98\textwidth,
    fontadjust=true,
    numbers=none,
    aboveskip=0.8\baselineskip,
    belowskip=0.8\baselineskip,
    columns=flexible,
    upquote=true,
    inputencoding=utf8,
    extendedchars=true,
    lineskip=-0.1pt,
    resetmargins=true,
    moredelim=[s][\color{blue!70!black}]{\{}{\}},
}

\lstdefinestyle{heuristicstyle}{
    backgroundcolor=\color{gray!5},
    commentstyle=\color{green!60!black},
    keywordstyle=\color{blue!70!black}\bfseries,
    numberstyle=\tiny\color{gray!70},
    stringstyle=\color{purple!70!black},
    basicstyle=\scriptsize\ttfamily\color{black},
    breakatwhitespace=false,
    breaklines=true,
    captionpos=b,
    keepspaces=true,
    showspaces=false,
    showstringspaces=false,
    showtabs=false,
    tabsize=4,
    frame=single,
    rulecolor=\color{gray!50},
    framesep=3pt,
    frameround=tttt,
    numbersep=6pt,
    xleftmargin=10pt,
    xrightmargin=10pt,
    aboveskip=1.0\baselineskip,
    belowskip=1.0\baselineskip,
    upquote=true,
    columns=flexible,
    keepspaces=true,
    mathescape=true,
    escapeinside={(*@}{@*)},
    morecomment=[l]\#,
    morekeywords={import, from, as, def, class, return, yield, for, while, if, elif, else, try, except, finally, with, lambda, pass, break, continue, and, or, not, is, in, raise, assert},
    emph={self, None, True, False, np, pd, plt, torch, tf, sklearn},
    emphstyle=\color{orange!80!black}\bfseries,
    literate=
        {-}{-}1
        {=>}{$\Rightarrow$ }3
        {->}{$\rightarrow$ }3
        {...}{$\ldots$ }3,
    inputencoding=utf8,
    extendedchars=true,
    lineskip=-0.1pt,
    fontadjust=true
}

\title{\our{}: Learning a Unified Solver for Profiled Routing}

\author {
    Chuanbo Hua\textsuperscript{\rm *, 1, 2},
    Federico Berto\textsuperscript{\rm *, 1, 2},
    Zhikai Zhao\textsuperscript{\rm 1},\\
    Jiwoo Son\textsuperscript{\rm 2},
    Changhyun Kwon\textsuperscript{\rm 1, 2},
    Jinkyoo Park\textsuperscript{\rm 1, 2}
}
\affiliations {
    \textsuperscript{\rm 1}KAIST, 
    \textsuperscript{\rm 2}OMELET, 
    AI4CO\textsuperscript{$\dagger$}
}

\begin{document}

\maketitle

\begin{abstract}
The Profiled Vehicle Routing Problem (PVRP) extends the classical VRP by incorporating vehicle–client-specific preferences and constraints, reflecting real‑world requirements such as zone restrictions and service‑level preferences. While recent reinforcement‑learning solvers have shown promising performance, they require retraining for each new profile distribution, suffer from poor representation ability, and struggle to generalize to out‑of‑distribution instances. In this paper, we address these limitations by introducing \textbf{U}nified \textbf{S}olver for \textbf{P}rofiled \textbf{R}outing (\our{}), a novel framework that natively handles arbitrary profile types. \our{} introduces on three key innovations: (i) {Profile Embeddings (PE)} to encode any combination of profile types; (ii) {Multi‑Head Profiled Attention (MHPA)}, an attention mechanism that models rich interactions between vehicles and clients;  (iii) {Profile‑aware Score Reshaping (PSR)}, which dynamically adjusts decoder logits using profile scores to improve generalization. Empirical results on diverse PVRP benchmarks demonstrate that \our{} achieves state‑of‑the‑art results among learning‑based methods while offering significant gains in flexibility and computational efficiency. We make our source code publicly available to foster future research.
\end{abstract}

\section{Introduction}
\label{sec:introduction}
\begin{figure}[t]
    \centering
    \includegraphics[width=\linewidth]{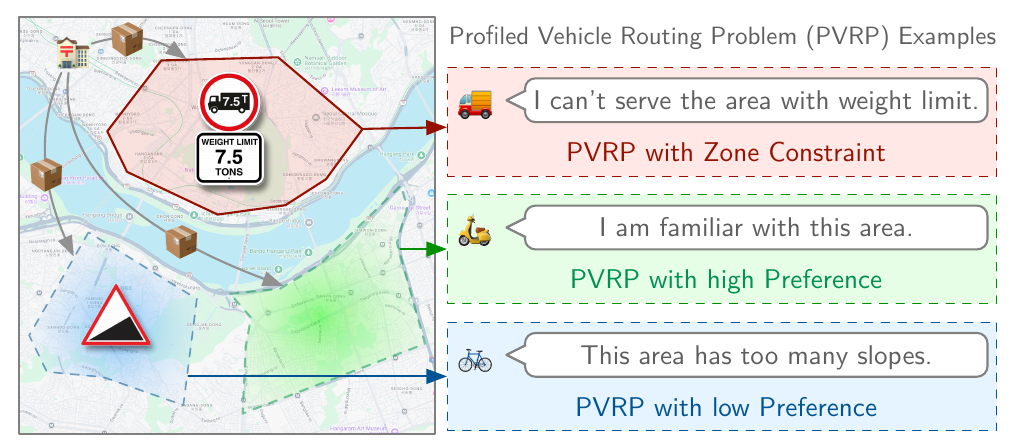}
    \caption{An illustrative example of the PVRP. The profiles are derived from real-world settings. Intuitively, zone constraints represent impassable regions; preference values indicate the ``desirability'' of an area for agents to visit.}
    \label{fig:catchy-figure}
    \vspace{-4mm}
\end{figure}

The Vehicle Routing Problem (VRP) is an important combinatorial optimization problem that focuses on determining optimal delivery routes for a fleet of vehicles serving a set of clients. In real-world logistics operations, vehicles often have distinct characteristics that affect their suitability for serving specific clients, leading to the Profiled Vehicle Routing Problem (PVRP). This variant extends traditional VRP constraints by incorporating vehicle-client-specific preferences and operational requirements \citep{Cordeau2001,braekers2016vehicle, zhong2007territory, aiko2018incorporating}. These profiles can represent various practical considerations: specialized vehicle access permissions in urban areas, client-specific service level agreements, regulatory restrictions, or historical performance metrics that influence routing decisions \citep{locus2020zonebased, li2023experience}. For instance, in last-mile delivery scenarios, certain vehicles might be preferred for specific neighborhoods based on size restrictions or noise regulations, while in B2B logistics, particular vehicle-driver combinations might maintain stronger relationships with certain clients. \cref{fig:catchy-figure} provides an example illustration for the PVRP problem. The PVRP's inherent complexity stems from its NP-hard nature, as it generalizes the classical VRP while adding profile-specific constraints that exponentially increase the solution space \citep{papadimitriou1998combinatorial, golden1984fleet}. This complexity becomes particularly challenging in modern logistics operations, where organizations must optimize routes for large fleets while considering numerous client-specific requirements and dynamically changing preferences.
Traditional approaches to solving PVRP typically rely on exact methods like Branch and Bound for small instances or metaheuristic algorithms such as genetic algorithms and simulated annealing for larger problems \citep{johnson1997traveling, lozano2011editorial}. While these methods can provide near-optimal solutions, they often require significant computational resources and extensive parameter tuning. Moreover, these approaches generally need to be redesigned or substantially modified when problem specifications change, such as when new types of preferences or constraints are introduced. This lack of adaptability poses a significant challenge in dynamic business environments where routing requirements frequently evolve.

Recent advances in neural combinatorial optimization (NCO), particularly through reinforcement learning (RL), have shown promising results for various VRP variants \citep{bello2016neural, sun2019fast}. These approaches leverage neural architectures, especially the pointer network paradigm \citep{vinyals2015pointer, kool2018attention, duanEfficientlySolvingPractical2020}, to learn solution strategies through interaction with simulated environments. While initial work focused on basic VRP variants \citep{kwon2020pomo, kim2022sym, zheng2024dpn}, recent studies have extended these methods to handle more complex scenarios, including multi-agent routing \citep{zong2022mapdp}, rich constrained variants \citep{liu2024multi, zhou2024mvmoe,bi2024learning}, and heterogeneous fleet problems extensible to the PVRP \citep{li2022deep, liu20242d, hua2025camp_vrp}.

However, existing learning‐based PVRP solvers exhibit three major shortcomings: (i) \textit{un-unified}, they must be retrained from scratch whenever the profile distribution or preference weights change, incurring prohibitive computational overhead; (ii) \textit{context-agnostic representation}, they lack the representational capacity to capture complex and diverse vehicle–client interactions, leading to suboptimal embeddings and degraded solution quality; and (iii) \textit{poor generalization}, they generalize poorly to out‐of‐distribution instances, making them fragile in dynamic or unseen settings. Together, these issues undermine the flexibility and efficiency required for real‐world deployment.

To bridge these limitations, we introduce \textbf{U}nified \textbf{S}olver for \textbf{P}rofiled \textbf{R}outing (\our{}), a transformer‐based policy that addresses the three core shortcomings of existing methods: (i) \textit{Profile Embeddings} (PE) encode arbitrary combinations of profile attributes and global weight parameters, removing the need to retrain for each new profile distribution;  (ii) \textit{Multi‐Head Profiled Attention} (MHPA) overcomes weak representation by capturing rich, bidirectional vehicle–client interactions; and (iii) \textit{Profile‐aware Score Reshaping} (PSR) adaptively reweights decoder logits with profile scores and spatial penalties, yielding robust generalization to out‐of‐distribution instances, especially to large-scale instances. We summarize our main contributions as follows:
\begin{itemize}
    \item We propose \our{}, the first unified neural solver for PVRP. A single \our{} model can generalize to diverse profile distributions without requiring retraining.
    
    \item We design the novel architecture built on three key components: PE for zero-shot adaptation to new profile distributions, MHPA to enhance representational capacity, and PAR to ensure robust generalization.
    
    \item We demonstrate through extensive experiments that a single \our{} model significantly outperforms existing state-of-the-art methods in solution quality and improvement on out-of-distribution real-world large-scale instances, while reducing both model size and training time by an order of magnitude.
\end{itemize}

\section{Related Work}
\label{sec:related_work}
\paragraph{Neural Combinatorial Optimization}
NCO has emerged as a powerful paradigm for solving VRP, offering promising end-to-end solutions that reduce the need for manual algorithm design \citep{bengio2021_ml4co_survey,berto2023rl4co,mazyavkina2021reinforcement,li2025unify}. The field was pioneered by \citet{vinyals2015pointer,bello2016neural} with Pointer Networks. These methods were significantly advanced by \citet{kool2018attention}'s seminal work, which introduced a transformer-based architecture trained via RL for solving VRPs, which remains the de facto basis for most modern neural VRP approaches. Recent developments in NCO for VRPs can be broadly categorized into construction and improvement methods. Construction methods \citep{kim2022sym, bogyrbayeva2023deep, grinsztajn2023winner,pirnay2024take,zhang2025adversarial} focus on generating solutions from scratch, while improvement/search methods \citep{hottung2020neural,li2021learning,li2024distribution, ma2024learning,ouyang2025learning} iteratively refine existing solutions. Construction approaches have seen significant innovations, including non-autoregressive methods \citep{kool2022deep, sun2024difusco} that predict promising edges simultaneously, and population-based approaches \citep{grinsztajn2024winner, hottung2024polynet} that maintain solution diversity. These have been complemented by advances in training strategies, such as problem re-encoding \citep{bdeir2022attention, drakulic2024bq} and test-time adaptation \& search \citep{hottung2021efficient, choo2022simulation,kim2025neural,kim2024gfacs,ye2023deepaco,hottung2025neural}. In this work, we focus on construction approaches for VRPs because of their adaptability to various settings and advantageous solution quality and inference time tradeoff. 

\paragraph{NCO for Practical VRPs}
As NCO methods mature, there is increasing focus on addressing ``in the wild'' VRPs -- VRP variants with complex constraints and real-world desiderata that can be applied in practical scenarios. \citet{duanEfficientlySolvingPractical2020, son2025neural} explore the gap between synthetic Euclidean and real-world asymmetric topological settings by modeling data distributions. Several works have extended to multiple complex constraints \citep{bi2024learning,liu2024multi} with several multi-task learning methods \citep{drakulic2024goal,liu2024multi,zhou2024mvmoe,berto2024routefinder,li2024cada,gohshield}. An important practical direction to model multiple vehicles in restricted numbers -- a realistic setting which most previous approaches do not consider --
has tackled multi-agent scenarios via multi-agent RL \citep{zong2022mapdp} and one-agent-at-a-time autoregressive reformulations \citep{son2024equity, zheng2024dpn}.  Some recent works tackle the setting of both limited vehicles and heterogeneous fleets modeling different vehicles  \citep{li2022deep, berto2024parco}, which are recently extended to handle the more practical PRVPs \citep{hua2025camp_vrp} that models not only different vehicle entities but also different vehicle-node interactions, i.e. \emph{profiles}, in terms of preferences of varying magnitudes and zone constraints. Despite recent progress, neural VRP solvers still struggle with weak profile modeling, require retraining for each new preference or constraint, and generalize poorly to unseen scenarios, which this work addresses.

\section{Preliminaries}

We introduce the problem formulation of the PVRP in a \textit{unified} manner in this section, including its Markov Decision Process (MDP) equivalent and the policy parametrization.

\subsection{Problem Formulation}

We consider a directed graph $G=(V,E)$, where
$V = \{0, \dots, N\}$ is the set of nodes, including $\{0\}$ as the depot and $\{1, \dots, N\}$ as clients, and vehicle set
$K = \{1,\dots,M\}$ is the set of vehicles. Each client
$i\in V$ has demand $d_i$, and each vehicle $k\in K$ has capacity $Q_k$,
speed $v_k$. Between each client $i$ and vehicle $k$, there is a profile score $p_{ik}\in\mathbb R\cup\{\pm\infty\}$. Intuitively, a higher profile score means that vehicle $k$ is encouraged to serve client $i$; symmetrically, this could also be understood as client $i$ preferring vehicle $k$ to serve them. Particularly, if the profile score is $\pm\infty$, it means a hard constraint, i.e., $-\infty$ means that the vehicle $k$ can not serve the client $i$, while $+\infty$ means that the vehicle $k$ has to serve the client $i$. The edges $E$ connect pairs of nodes, and each edge between node $i$ and node $j$, $(i, j) \in E$, has a travel distance $c_{ij}$. We introduce the decision variables
$x^k_{ij}=1$ if vehicle $k$ travels from $i$ to $j$ (and $0$ otherwise)
and $y^k_i=1$ if vehicle $k$ serves client $i$ (and $0$ otherwise).
The objective is to maximize total profile reward minus travel time, balanced via a \textit{profile weight} $\alpha \in [0, 1]$:
\begin{equation}
\label{eq:objective}
\max_{x,y}\;\sum_{k\in K}\sum_{i\in V}\sum_{j\in V}
\bigl(\alpha\,p_{ik}-\tfrac{c_{ij}}{v_k}\bigr)\,x^k_{ij}.
\end{equation}
under the constraints that each client is served exactly once, vehicle capacities are not exceeded, each route is a continuous tour beginning and ending at the depot, and all decision variables are binary. By setting all $p_{ik}=0$, we recover the classical VRP\footnote{Specifically, this case would be a Capacitated Vehicle Routing Problem (CVRP) where the objective is to minimize the total travel time (duration).}. A more detailed problem formulation of PVRP is provided in the Appendix.

\subsection{MDP Formulation}
\label{sec:mdp-formulation}
The PVRP can be naturally framed as a MDP, which enables the application of RL techniques for scalable and adaptive solution generation. We define the formulation as follows:

\paragraph{State Space ($\mathcal{S}$)} A state $s^t \in \mathcal{S}$ at time step $t$ captures the partial route constructed up to that point.

\paragraph{Action Space ($\mathcal{A}$)} An action $a\in\mathcal{A}$ consists of selecting the client to visit next or returning to the depot. 

\paragraph{Transition Dynamics ($\mathcal{T}$)} The system evolves according to a deterministic transition function $s_{t+1} = \mathcal{T}(s_t, a_t)$, which updates vehicle locations, remaining capacities, and the set of visited clients based on the selected action. 

\paragraph{Reward Function ($\mathcal{R}$)} To align with the \textit{bi-objective} nature of PVRP, we define the reward for each action as $r_t = \alpha p_{jk} - c_{ij}/v_k$ as per \cref{eq:objective}, where $p_{jk}$ represents the client-vehicle preference score, $c_{ij}/v_k$ accounts for the travel cost, and $\alpha$ controls the trade-off between preference satisfaction and transportation efficiency. 

\paragraph{Policy ($\pi$)} A policy $\pi(a_t | s_t)$ specifies the probability distribution over possible actions given the current state. Our objective is to learn an optimal policy $\pi^*$ that maximizes the expected cumulative reward.

\subsection{Policy Parameterization}
\label{sec:policy-parametrization}
To generate solutions efficiently, we employ parallel autoregressive models for policy learning with an encoder-decoder framework. The encoder network $f_\theta(\bm{x}, \alpha)$ processes problem instance $\bm{x}$ and \textit{profile weight} $\alpha\in[0,1]$ into a structured representation $\bm{h}$. At each step $t$, the decoder network $g_\theta$ generates a joint action vector $\bm{a}_t = (a_t^1, \ldots, a_t^M)$ for all $M$ vehicles. The policy $\pi_\theta$ is formulated as:
\begin{equation*}
\pi_\theta(\bm{a} | \bm{x}, \alpha) = \prod_{t=1}^{T} \psi\left(\prod_{k=1}^{M} g_\theta(a_{t}^k | \bm{a}_{t-1}, \bm{a}_{t-2}, \ldots, \bm{a}_1, \bm{h})\right)
\end{equation*}
where $\psi$ is a conflict resolution function that ensures solution feasibility by prioritizing assignments to the agent with the largest log-probability value.

\section{Methodology}
\label{sec:methodology}
\begin{figure*}[h!]
    \centering
    \includegraphics[width=\linewidth]{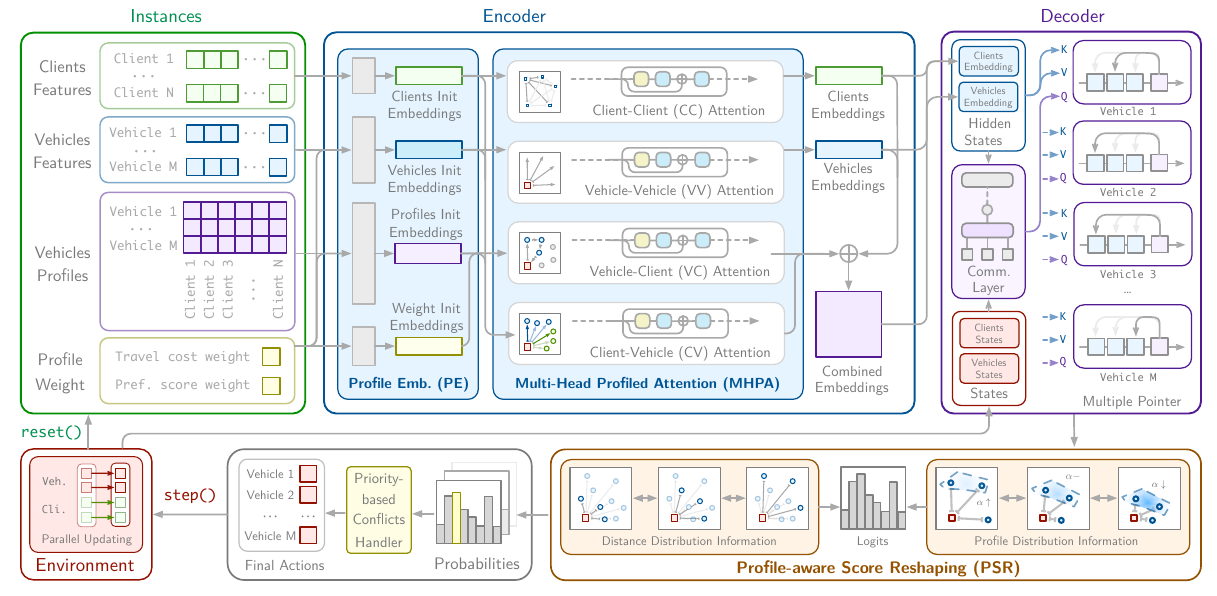}
    \vspace{-6mm}
    \caption{An overview of \our{}. Our framework follows an encoder-decoder architecture and introduces three key components to handle diverse problem profiles within a single model. (i) \textbf{PE} firstly encode arbitrary instance inputs, including client-vehicle profiles and \textit{profile weights}, into a unified latent representation. (ii) \textbf{MHPA} then captures rich, bidirectional interactions between all entities to produce powerful, context-aware embeddings. (iii) \textbf{PSR} adaptively integrates profile and distance distribution and scale information to adjust the policy output logits, ensuring the robust generalization ability.}
    \label{fig:overview}
    \vspace{-3mm}
\end{figure*}
In this section, we present \our{} as illustrated in \cref{fig:overview}. We introduce three key components for \textit{unified} profile handling: \textit{profile embeddings} (PE),  \textit{multi-head profiled attention} (MHPA), and \textit{profile-aware score reshaping} (PSR). Together, these components enable flexible adaptation to different profile distributions and problem settings within a single model architecture. We then lay out the integration into the encoder-decoder framework and the training scheme. 

\subsection{Profile Embeddings}
\label{sec:pe}
Previous approaches handle profiles by \textit{retraining} separate models for each profile distribution and profile weight, which is computationally inefficient and lacks flexibility. PE overcome this limitation by learning a unified representation that can encode any combination of attributes and profile weight parameters. Given an instance $\bm{x}$ and a profile weight parameter $\alpha$, we first embed these features into a latent space $\mathbf{h}^{(\cdot)}$ of size $d_h$ by considering separated contributions for clients, vehicles, profiles, and profile weight with the dimension of $d_{(\cdot)}$ via parametrized linear layers.

\paragraph{Client Feature Embeddings} project each client $i$ client-specific information to the hidden space: $\mathbf{h}^c_i = \mathbf{W}^c_{\text{init}}\bm{x}^c_i + \mathbf{b}^c_{\text{init}} \in \mathbb{R}^{d_h}$, where $\bm{x}^c_i\in\mathbb{R}^{d_c}$ contains demands and locations, here $\mathbf{W}^c_{\text{init}} \in \mathbb{R}^{d_h \times d_c},\mathbf{b}^c_{\text{init}} \in \mathbb{R}^{d_h}$.

\paragraph{Vehicle Feature Embeddings} project each vehicle $k$ vehicle-specific information to the hidden space: $\mathbf{h}^v_k = \mathbf{W}^v_{\text{init}}\bm{x}^v_k + \mathbf{b}^v_{\text{init}} \in \mathbb{R}^{d_h}$, where $\bm{x}^v_j\in\mathbb{R}^{d_v}$ contains capacity, speed, and starting location, which is practically the depot information, here $\mathbf{W}^v_{\text{init}} \in \mathbb{R}^{d_h \times d_v}, \mathbf{b}^v_{\text{init}} \in \mathbb{R}^{d_h}$.

\paragraph{Profiles Score Embeddings} transform raw profile matrix into learnable embeddings:
$\mathbf{h}^p_{ik} = \mathbf{W}^p_{\text{init}}p_{ik} + \mathbf{b}^p_{\text{init}} \in \mathbb{R}^{d_h}$, where $p_{ik}\in\mathbb{R}$ represents the profile score between vehicle $i$ and client $k$, $\mathbf{W}^p_{\text{init}} \in \mathbb{R}^{d_h \times 1}$, and $\mathbf{b}^p_{\text{init}} \in \mathbb{R}^{d_h}$. For a hard constraint profile score, we have the embeddings by projecting with independent parametrized linear layers: $\mathbf{h}^{p}_{ik} = \mathbf{W}^{p,\pm\infty}_{\text{init}}\mathbf{1} + \mathbf{b}^{p,\pm\infty}_{\text{init}}$. With this design, our model could flexibly embed any type of mixed combination of soft-preferenced and hard-constrained profile matrix.  

\paragraph{Profile Weight Embeddings} encode the weight to enable flexible tradeoffs:
$\mathbf{h}^{\alpha} = \mathbf{W}^{\alpha}_{\text{init}}\mathbf{\alpha} + \mathbf{b}^{\alpha}_{\text{init}} \in \mathbb{R}^{d_h}$, where $\alpha \in \mathbb{R}$ represents the profile weights from \cref{eq:objective} which are then broadcasted on other embeddings, $\mathbf{W}^{\alpha}_{\text{init}} \in \mathbb{R}^{d_h \times 1}$, and $\mathbf{b}^{\alpha}_{\text{init}} \in \mathbb{R}^{d_h}$. This simple yet effective design allows the model to dynamically adapt to various objective weights.

PE create a shared latent space that enables the model to handle arbitrary profile types and profile weights without retraining. By projecting the client, vehicle, profiles, and profile weights, the model can simultaneously process profiles with varying magnitudes and values.

\subsection{Multi-Head Profiled Attention}
\label{sec:mhpa}
A fundamental limitation of existing approaches is their inability to capture rich bidirectional interactions between vehicles and clients with different profiles. To address this, we introduce MHPA to allow for improved information exchange in each encoder layer:
\begin{align}
    \mathbf{h} &= \text{Norm}(\text{MHPA}(\mathbf{h})) \\
    \mathbf{h} &= \text{Norm}(\text{FFN}(\mathbf{h}) + \mathbf{h})
\end{align}
where $\text{FFN}(\cdot)$ denotes a multi-layer perceptron. MHPA is based on the multi-head attention (MHA):
\begin{align*}
    &\text{MHA}(\mathbf{Q},\mathbf{K},\mathbf{V}) = \left( \mathop{\big\|}_{i=1}^{n_\text{h}} \text{Attn}(\mathbf{Q}\mathbf{W}^Q_i, \mathbf{K}\mathbf{W}^K_i, \mathbf{V}\mathbf{W}^V_i) \right) \mathbf{W}^O\\
    &\text{Attn}(\mathbf{Q}, \mathbf{K}, \mathbf{V}) = \text{softmax} \left( \frac{\mathbf{Q}\mathbf{K}^\top}{\sqrt{d_h}} \right) \mathbf{V}
\end{align*}

MHPA improves on MHA for modeling PVRP with four distinct types of information exchange between clients and vehicles bi-directionally, depending on the init embedding:

\paragraph{Client-Client (CC) Attention} enables information sharing between clients about their spatial location, distance relationships, and demands: $\mathbf{h}^{c'} = \text{MHA}(\mathbf{h}^{c}, \mathbf{h}^{c}, \mathbf{h}^{c})$.
\paragraph{Vehicle-Vehicle (VV) Attention} facilitates communication between vehicles about their current locations, speed, distance relationships, maximum and available capacities, and service capabilities: $\mathbf{h}^{v'} = \text{MHA}(\mathbf{h}^{v}, \mathbf{h}^{v}, \mathbf{h}^{v})$.
\vspace{-1mm}
\paragraph{Vehicle-Client (VC) Attention} enables vehicles to attend to relevant clients based on \textit{profiles} about soft preferences or hard constraints: $\mathbf{h}^{pc'}_{i} = \text{MHA}(\mathbf{h}^{v}, \mathbf{h}^{c}, \mathbf{h}^{pv}_{i})$.
\vspace{-1mm}
\paragraph{Client-Vehicle (CV) Attention} allows clients to consider their suitability for different vehicles based on their \textit{profiles} in a reversed way compared with the VC attention to enrich the relationship embeddings: $\mathbf{h}^{pv'}_{k} = \text{MHA}(\mathbf{h}^{c}, \mathbf{h}^{v}, \mathbf{h}^{pc}_{k})$.

\vspace{2mm}
Unlike previous approaches that only consider unidirectional interactions or simple aggregations, MHPA enables comprehensive information exchange through bidirectional attention mechanisms. The final MHPA output integrates all processed information:
\vspace{-1mm}
\begin{equation}
    \mathbf{h}^{p'}_{ik} = \text{concat}(\mathbf{h}^{v}_k, \mathbf{h}^{c}_i, \mathbf{h}^{pv'}_{k}, \mathbf{h}^{pc'}_{i}) + \mathbf{h}^{p}_{ik}
\end{equation}
where each vehicle $k$ is assigned its own hidden embeddings $\mathbf{h}$ processed latent representation. 

\subsection{Profile-Aware Score Reshaping}
\label{sec:psr}
As we are using one unified model for various distributions of profiles and scales, to maintain robustness and out-of-distribution performance, we further introduce PAR, which dynamically adjusts decoder logits based on the distance and profile distributions. Building on recent advancements in distance-based heuristics \citep{wang2024distance,huang2025rethinking}, PSR further combines learned embeddings with explicit profile and distance information:
\vspace{-1mm}
\begin{equation}
    \mathbf{Z} = C \cdot \text{tanh}\left(\frac{\mathbf{U}(\mathbf{W}_{\text{ptr}}\mathbf{h})^\top}{\sqrt{d_h}} - \log(\text{dist}_{ij} + p_{ij})\right)
\end{equation}
where $C$ is a scaling factor set to $10$ following \citet{bello2016neural}, $\mathbf{U} \in \mathbb{R}^{M \times d_h}$ represents the decoder's query vectors, $\mathbf{W}_{\text{ptr}}$ projects node embeddings into the pointer space, $\text{dist}_{ij}$ is the distance between nodes $i$ and $j$, and $p_{ij}$ is the preference score between vehicle $i$ and client $j$\footnote{We note that while we manually design the attention reshaping mechanism in this paper, we could leverage automatic algorithm design \citep{liu2024evolution,ye2024reevo,pham2025hsevo,tranlarge,zhao2025trajevo} to do it, which we leave as future work.}.

PSR provides several key advantages: it balances the flexibility of neural approaches with the reliability of traditional heuristics; naturally penalizes nodes that are either spatially distant or have low preference scores; ensures consistent performance across varying profile distributions and out-of-distribution instances; and enables smooth adaptation without retraining when preference weights change. The final action probabilities are computed by masking infeasible actions and applying softmax:
\begin{equation}
    P(\mathbf{a}_t | \mathbf{s}_t) = \text{softmax}(\mathbf{Z} + \mathbf{M}_t)
\end{equation}
where $\mathbf{M}_t$ is the mask tensor for infeasible actions at step $t$.

\subsection{Integration into Encoder-Decoder Framework}
\label{sec:framework-integration}
The three components described above are integrated into an encoder-decoder framework. The encoder first processes raw problem features using PE to create initial embeddings, then applies encoder layers with MHPA to capture complex interactions between clients and vehicles. The encoder outputs a set of embeddings $\mathbf{h} = [\mathbf{h}^{1}, \ldots, \mathbf{h}^{M}]$, where each $\mathbf{h}^k \in \mathbb{R}^{(M+N) \times d_h}$ represents the encoded graph information for vehicle $k$. The decoder generates vehicle-specific queries that capture both profile and current state info:
\vspace{-1mm}
\begin{equation}
    \mathbf{q}^t_k = \mathbf{W}_{\text{query}}[\mathbf{h}^k \| \mathbf{h}^k_{\text{cur}} \| \mathbf{W}_{\text{state}}\mathbf{s}^t_k]
\end{equation}
where $\cdot\|\cdot$ denotes concatenation, $\mathbf{h}^k$ is the vehicle's profile embedding, $\mathbf{h}^k_{\text{cur}}$ represents the current node embedding, and $\mathbf{s}^t_k$ captures the state features at time $t$. We then process the multiple vehicle queries $\mathbf{q}_t = [\mathbf{q}_t^1, \dots, \mathbf{q}_t^M]$ via a communication layer. The communication layer follows \citet{berto2024parco} and is composed of standard MHA and FFN: this processes embeddings into a multiple pointer mechanism that generates the decoder output logits. Finally, PSR is applied to reshape the logits and compute action probabilities for each vehicle as the action.

\subsection{Training}
\label{sec:training}
We train our model using the REINFORCE algorithm with a shared baseline across all agents \citep{kim2022neuro}. During training, we sample weights $\alpha_i$ from $[0,1]$ and $p_{ik}$ from $[p_{\text{min}}, p_{\text{max}}]$ with a predefined probability to be $\pm\infty$ for each instance in the batch. This allows the model to learn a unified policy across different preference-cost trade-offs and profile matrix distributions. The policy gradient is estimated as:
\vspace{-2mm}
\begin{equation}
    \label{eq:reinforce}
    \begin{split}
        \nabla_\theta \mathcal{L} = \frac{1}{B \cdot L} \sum_{i=1}^{B} \sum_{j=1}^{L} 
        &(R(\bm{x}_i, \bm{a}_{ij}, \alpha_i) - b^{\text{shared}}(\bm{x}_i)) \\
        &\cdot \nabla_\theta \log p_\theta(\bm{a}_{ij} | \bm{x}_i, \alpha_i)
    \end{split}
\end{equation}
where $B$ is the batch size, $L$ is the number of solutions per instance, and $b^{\text{shared}}$ is the shared baseline value obtained through symmetric augmentation sampling.

\section{Experiments}
\label{sec:experiments}
We evaluate the effectiveness of our proposed approach with comprehensive experiments in various settings, including in-distribution performance analysis, large-scale (out-of-distribution) generalization analysis, real-world application, model components ablation study, and further analyses. We compare against both classical and learning-based baselines.

\subsection{Experimental Setup}
\label{sec:experimental-setup}

\paragraph{Data Generation}
For basic features of VRP, including the clients and depot coordinates, demands, capacity, and speed, we follow the widely used settings from \citep{kool2018attention}. Profile scores are drawn as $\sim\mathrm{Uniform}(0,1)$ independently for all client–vehicle pairs. Without losing the generalization, we sample two probabilities $\sim\mathrm{Uniform}(0,0.1)$ for each instance to randomly set part of the profile matrix to $\pm\infty$ as the hard constraints.

\paragraph{Classical Baselines}
We employ two state-of-the-art classical solvers: Google OR-Tools \citep{perron2023ortools}, a versatile framework that combines exact and heuristic methods via constraint programming, and HGS-PyVRP \citep{wouda2024pyvrp}, an open-source implementation of the Hybrid Genetic Search for the CVRP \citep{vidal2022hybrid} that supports the PVRP. We handle vehicle-specific profiles by modifying the cost matrices for each vehicle according to the objective function in \cref{eq:objective} and masking for hard constraints.

\begin{table*}[t!]
\small
\centering
    \begin{tabular}{l| c c c c | c c c c | c c c c | c}
    \toprule
    $N$ & \multicolumn{4}{c|}{60} & \multicolumn{4}{c|}{80} & \multicolumn{4}{c|}{100} & Gap($\%$)  \\
    \midrule
    $M$ & 3 & 5 & 7 & Time & 3 & 5 & 7 & Time & 3 & 5 & 7 & Time & avg.\\
    \midrule
    OR-Tools & 7.98 & 8.25 & 8.45 & 10m & 9.33 & 9.86 & 10.02 & 12m & 11.25 & 11.41 & 11.67 & 15m & 10.24 \\ 
    HGS-PyVRP & 7.07 & 7.42 & 7.66 & 10m & 8.51 & 8.97 & 9.23 & 12m & 9.99 & 10.51 & 10.78 & 15m & 0.00 \\ 
    \midrule
    ET (\textit{g.}) & 8.31 & 8.77 & 9.02 & 0.17s & 9.98 & 10.50 & 10.79 & 0.23s & 11.69 & 12.28 & 12.70 & 0.29s & 17.38 \\
    DPN (\textit{g.}) & 8.23 & 8.65 & 8.88 & 0.18s & 9.87 & 10.51 & 10.88 & 0.23s & 11.68 & 12.21 & 12.51 & 0.29s & 16.59 \\
    2D-Ptr (\textit{g.}) & 8.01 & 8.38 & 8.62 & 0.15s & 9.61 & 10.14 & 10.41 & 0.20s & 11.25 & 11.89 & 12.21 & 0.25s & 12.97 \\
    PARCO (\textit{g.}) & 7.98 & 8.36 & 8.66 & 0.15s & 9.59 & 10.04 & 10.37 & 0.22s & 11.26 & 11.85 & 12.12 & 0.25s & 12.62 \\
    CAMP (\textit{g.}) & 7.79 & 8.26 & 8.46 & 0.18s & 9.38 & 9.87 & 10.22 & 0.25s & 10.98 & 11.65 & 11.91 & 0.33s & 10.50\\
    \our{} (\textit{g.}) & \textbf{7.73} & \textbf{8.11} & \textbf{8.38} & 0.11s & \textbf{9.30} & \textbf{9.78} & \textbf{10.08} & 0.14s & \textbf{10.90} & \textbf{11.44} & \textbf{11.76} & 0.20s & \textbf{9.20} \\
    \midrule
    ET (\textit{s.}) & 7.82 & 8.23 & 8.45 & 0.25s & 9.40 & 9.89 & 10.20 & 0.36s & 11.06 & 11.64 & 11.90 & 0.45s & 10.59 \\
    DPN (\textit{s.}) & 7.79 & 8.18 & 8.46 & 0.26s & 9.36 & 9.88 & 10.18 & 0.36s & 10.99 & 11.58 & 11.92 & 0.44s & 10.25 \\
    2D-Ptr (\textit{s.}) & 7.55 & 7.93 & 8.17 & 0.17s & 9.12 & 9.55 & 9.87 & 0.21s & 10.69 & 11.19 & 11.49 & 0.26s & 6.79 \\
    PARCO (\textit{s.}) & 7.53 & 7.86 & 8.13 & 0.22s & 9.03 & 9.52 & 9.82 & 0.34s & 10.56 & 11.18 & 11.51 & 0.41s & 6.25 \\
    CAMP (\textit{s.}) & 7.39 & 7.77 & 8.01 & 0.34s & 8.90 & 9.37 & 9.67 & 0.42s & 10.44 & 10.98 & 11.27 & 0.53s & 4.58 \\
    \our{} (\textit{s.}) & \textbf{7.35} & \textbf{7.71} & \textbf{7.96} & 0.22s & \textbf{8.85} & \textbf{9.32} & \textbf{9.60} & 0.33s & \textbf{10.39} & \textbf{10.93} & \textbf{11.22} & 0.42s & \textbf{4.02} \\
    \bottomrule
    \end{tabular}
\caption{Benchmarks and results for PVRP at varying sizes and agent numbers. Highlighting cost ($\downarrow$) and average (avg.) gaps ($\downarrow$) to the HGS-PyVRP solver. The average inference time for a single instance of each size is shown in the Time columns.}
\label{table:main}
\vspace{-3mm}
\end{table*}

\paragraph{Neural Baselines}
We compare against several recent neural VRP solvers: ET \citep{son2024equity}, which specializes in sequential multi-agent routing with equitable workload distribution; DPN \citep{zheng2024dpn}, which enhances ET with an improved encoder for route partitioning; 2D-Ptr \citep{liu20242d}, which uses dual encoding for dynamic adaptation in heterogeneous routing; PARCO \citep{berto2024parco}, which employs parallel decoding with inter-agent communication; and CAMP \citep{hua2025camp_vrp}, which was specifically designed for PVRP. We follow CAMP to adapt ET, DPN, 2D-Ptr, and PARCO to PVRP for a fair comparison. 

\paragraph{Training Configuration} We optimize using Adam \citep{kingma2014adam} with an initial learning rate of $10^{-4}$, decaying by a factor of 0.1 at epochs 80 and 95. Training runs for 100 epochs with $10^5$ samples per epoch, using a batch size of 32 and 8 augmented rollouts via Sym-NCO \citep{kim2022sym} per instance for baseline estimation. Architecturally, each model employs $d_h=128$ hidden‐dimensional embeddings, 8 attention heads, and 512‐dimensional feedforward layers across 3 encoder layers. We set the scale of the number of agents and vehicles during training from $60$ to $100$, from $3$ to $7$, respectively. For baseline models, following the setting from CAMP, we train separate models for each fixed $\alpha$ taken from $\alpha \in \{0.00, 0.025, \ldots, 0.20\}$, where the final reported performance is the average of results across all $\alpha$ values. As a unified model, \our{} is trained with the $\alpha$ is randomly sampled from $0$ to $0.2$ for each instance, also including a sampling probability for the mixed hard-constraints from $0$ to $10\%$. All training and testing are run on an AMD Ryzen Threadripper 3960X (24-core) CPU with an NVIDIA RTX 4090 GPU. For more detailed training hyperparameters and device information, please refer to the Appendix.

\paragraph{Testing Protocol}
We consider three main settings for evaluating our approach. Firstly, we consider in-distribution results, where we evaluate each model on 1,280 randomly generated instances for each size, following the same generating rule as training. Secondly, we test 128 instances in out-of-distribution generalization, with vehicle numbers $M\in\{15,25,35\}$ for $N=500$. We finally introduce a variation of the real-world data of CVRPLib\footnote{\url{http://vrp.atd-lab.inf.puc-rio.br/index.php/en/}}, which we coin \textit{PVRPLib}, which is based on the number of vehicles, capacity values, and coordinates of the original CVRPLib but with profiles generated as described in the data generation paragraph. For each instance, we measure both greedy performance ($g.$), i.e., taking the \(\arg\max\) over decoder log‐probabilities, and sampling 1,280 solutions per instance ($s.$). Final performance metrics are reported as averages over all profile distributions and \(\alpha\) settings.

\subsection{Main Results and Analysis}
We address the three limitations of prior works in the following paragraphs: (i) the lack of a \textit{unified model}, (ii) \textit{context-agnostic representations}, and (iii) \textit{poor generalization}. 

\paragraph{Unified Model}
A primary advantage of our unified model is its exceptional efficiency. As demonstrated in Table 2, our approach significantly reduces total training time compared to training multiple single-task models like CAMP. By using a single, shared architecture, we substantially lower both memory and computational costs, cutting the total number of required parameters. This streamlined process not only accelerates the training cycle but also maintains competitive performance. Consequently, our method eliminates the need to develop and manage separate models for various problem types, offering a scalable and adaptable solution for PVRP that ensures high-quality optimization across diverse vehicle-client profiles.

\begin{table}[h]
\small
\setlength{\tabcolsep}{1mm}
\centering
\label{tab:unified-model-time}
\begin{tabular}{lcccc}
\toprule
 & \# Models & \# Total Param. & \# Total Epochs & Train Time \\
\midrule
CAMP  & 10 & 17.6M  & 1000  & 4.6 days  \\ 
\our{}  & \textbf{1} & \textbf{1.5M}  & \textbf{100}  & \textbf{11 hours}  \\
\bottomrule
\end{tabular}
\caption{Our unified model enables substantial memory and training time savings compared to single-task CAMP.}
\vspace{-2mm}
\end{table}

\paragraph{In-distribution Performance}
\cref{table:main} shows a comparison between our method and the baseline models on the in-distribution scale. The results demonstrate that our \our{} achieves state-of-the-art performance for all problem settings, consistently outperforming all existing neural solvers in both solution quality and computational efficiency, as well as outperforming Google OR-Tools while at a fraction of the computational cost. Note that, unlike previous approaches that require training separate models for different preference weights, \our{} is trained as a \textit{single} model, effectively handling varying client-vehicle constraints and preference distributions within a unified framework. 

\begin{table}[h!]
\centering
\small
\setlength{\tabcolsep}{1mm}
\begin{tabular}{l ccc c l}
\toprule
M & 30 & 50 & 70 & Time & Gap(\%) \\
\midrule
OR-Tools & 102.86 & 112.65 & 115.96 & 15m & 58.98 \\
HGS-PyVRP & 64.70 & 70.86 & 72.94 & 15m & 0.00 \\
\midrule
ET (\textit{g.}) & 123.73 & 135.51 & 139.48 & 2.34s & 91.23 (+71.88)\\
DPN (\textit{g.}) & 109.58 & 120.01 & 123.53 & 2.34s & 69.36 (+50.01)\\
2D-Ptr (\textit{g.}) & 91.49 & 100.20 & 103.14 & 1.07s & 41.41 (+22.06)\\
PARCO (\textit{g.}) & 81.24 & 88.97 & 91.58 & 1.24s & 25.56 (+ \- 6.21)\\
CAMP (\textit{g.}) & 80.05 & 87.67 & 90.24 & 1.33s & 23.72 (+ \- 4.37) \\
\our{} (\textit{g.}) & \textbf{77.22} & \textbf{84.57} & \textbf{87.05} & \textbf{1.07s} & \textbf{19.35} \\
\midrule
ET (\textit{s.}) & 117.90 & 129.13 & 132.92 & 2.59s & 82.23 (+68.08)\\
DPN (\textit{s.}) & 100.69 & 110.28 & 113.52 & 2.58s & 55.63 (+41.48)\\
2D-Ptr (\textit{s.}) & 85.27 & 93.39 & 96.13 & 1.18s & 31.80 (+17.65)\\
PARCO (\textit{s.}) & 80.82 & 88.52 & 91.12 & 1.37s & 24.92 (+10.77)\\
CAMP (\textit{s.}) & 79.87 & 87.47 & 90.04 & 1.43s & 23.44 (+ \- 9.29)\\
\our{} (\textit{s.}) & \textbf{73.86} & \textbf{80.89} & \textbf{83.26} & \textbf{1.18s} & \textbf{14.15} \\
\bottomrule
\end{tabular}
\label{tab:largescale}
\caption{Benchmarks and results for large-scale PVRP instances ($N=1000$). We report the solution cost ($\downarrow$) and average gap ($\downarrow$) to the  HGS-PyVRP solver. Average inference time is shown in the Time column.}
\vspace{-3mm}
\end{table}

\paragraph{Out-of-Distribution Performance} 
Table 3 shows the out-of-distribution performance, particularly in large scales up to $10\times$ the number of agents $M$ and $10\times$ the number of nodes $N$. Our model outperforms all previous neural methods. This validates the advantage of a unified model on superior profile handling, robustness, and generalization ability. For more scaling results, please refer to the Appendix.

\begin{table}[h]
\small
\setlength{\tabcolsep}{1mm}
\centering
\label{tab:results-cvrplib-p}
\begin{tabular}{lcccccc}
\toprule
\multirow{2}{*}{} & \multirow{2}{*}{Size} & \multirow{2}{*}{BKS} & \multicolumn{2}{c}{CAMP} & \multicolumn{2}{c}{USPR} \\
\cmidrule(lr){4-5} \cmidrule(lr){6-7}
 &  &  & Cost & Gap & Cost & Gap (\%) \\
\midrule
Set A & 31-79 & 9.24 & 9.87 & 6.82(+1.63) & {9.72} & \textbf{5.19}  \\
Set B & 30-77 & 10.18 & 10.80 & 6.09(+0.88) & {10.71} & \textbf{5.21} \\
Set F & 44-134 & 12.68 & 13.58 & 7.10(+0.48) & {13.52} & \textbf{6.62} \\
Set M & 100-199 & 45.63 & 54.39 & 19.20(+5.74) & {51.77} & \textbf{13.46} \\
Set P & 15-100 & 9.09 & 9.67 & 6.38(+1.32) & {9.55} & \textbf{5.06} \\
\midrule
\multirow{4}{*}{Set X} & 100-300 & 15.56 & 17.43 & 12.01(+3.29) & {16.92} & \textbf{8.72} \\
 & 300-500 & 38.08 & 44.79 & 17.62(+5.96) & {42.52} & \textbf{11.66} \\
 & 500-700 & 66.04 & 78.42 & 18.74(+5.32) & {74.90} & \textbf{13.42} \\
 & 700-1K & 99.08 & 124.61 & 25.77(+10.56) & {114.15} & \textbf{15.21} \\
\bottomrule
\end{tabular}
\caption{Results of CAMP and \our{} about cost ($\downarrow$) and average gap ($\downarrow$) across PVRPLib instances. The best-known solutions (BKS) are collected by the HGS-PyVRP solver.}
\vspace{-3mm}
\end{table}

\paragraph{Real-world Settings} 
We further analyze the performance of \our{} against the SOTA neural method CAMP on the newly proposed PVRPLib, containing real-world location distributions, demands, number of vehicles, and the added preferences in Table 4. Our method remarkably improves on CAMP by more than $10\%$ in large-scale instances.

\begin{table}[h!]
\centering
\label{tab:ablation_horiz}
\small
\setlength{\tabcolsep}{1mm}
\begin{tabular}{lcccc}
\toprule
Model & USPR & -PSR & -PSR \& SR & -PSR, SR \& MHPA \\
\midrule
Avg. Gap (\%) & \textbf{4.42} & 4.72 & 5.12 & 6.72 \\
\bottomrule
\end{tabular}
\caption{Ablation study results on the size of $N=100$.}
\end{table}

\paragraph{Ablation Study} We perform an ablation study to evaluate the contribution of model components in Table 5. We remove the PSR, SR (distance only), and MHPA (same as baselines). Removing each component will cause a drop in performance, showing the importance of each design. The most significant drop occurs when eliminating the MHPA, highlighting its critical role in improving inter-agent communication and capturing profile-specific interactions. 

\paragraph{Qualitative Analysis}
Figure 3 visualizes example solutions for a PVRP instance. With increasing the \textit{profile weight} $\alpha$, our model shifts focus towards profile adherence while maintaining strong duration optimization, effectively capturing vehicle-client interactions. A single unified model is used to solve for any value of $\alpha$. It demonstrates how increasing its value makes the model trade off duration for overall preferences. This makes our model highly scalable and adaptable for real-world applications.

\begin{figure}[h!]
    \centering
    \includegraphics[width=\linewidth]{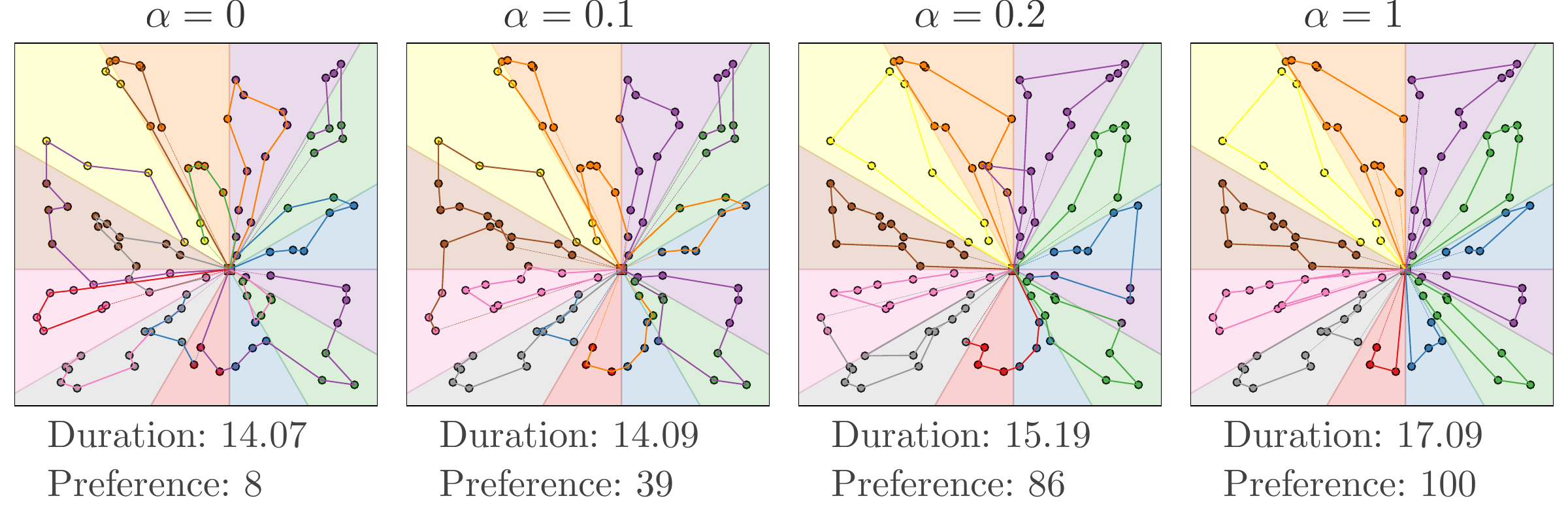}
    \caption{\our{}'s PVRP solutions to the same instance with different $\alpha$ processed through the profile embeddings. A higher $\alpha$ favors adherence to profile values, while $\alpha=0$ converts the problem to a classical VRP.} 
    \label{fig:visualization}
\end{figure}

\section{Conclusion}
\label{sec:conclusion}
In this work, we introduced \our{}, a learning-based framework that addresses the key limitations of existing PVRP neural solvers through three novel components: PE for encoding arbitrary profile distributions, MHPA for modeling rich vehicle-client interactions, and PSR for robust generalization. Our comprehensive experiments demonstrate that \our{} consistently outperforms state-of-the-art neural methods across both preference-based and zone-constrained routing problems, while matching or exceeding classical solvers at significantly lower computational cost. Notably, a single \our{} model effectively handles varying profile weights and generalizes to out-of-distribution instances up to 10$\times$ larger than training data. By providing this unified approach to profiled routing optimization and making our implementation publicly available, we aim to advance NCO research and enable more flexible, efficient solutions for complex routing problems in real-world logistics operations. A limitation to address in future works is reducing the memory usage of models. Both \our{} and baselines embed the profile in a matrix way, which will easily explode when the scale of the instance increases to an extremely large scale ($\geq 10,000$). 

\clearpage
\pagebreak 

\bibliography{aaai2026}

\begin{thebibliography}{69}
\providecommand{\natexlab}[1]{#1}

\bibitem[{Aiko, Thaithatukl, and Asakura(2018)}]{aiko2018incorporating}
Aiko, S.; Thaithatukl, P.; and Asakura, Y. 2018.
\newblock Incorporating user preference into optimal vehicle routing problem of integrated sharing transport system.
\newblock \emph{Asian Transport Studies}, 5(1): 98--116.

\bibitem[{Bdeir, Falkner, and Schmidt-Thieme(2022)}]{bdeir2022attention}
Bdeir, A.; Falkner, J.~K.; and Schmidt-Thieme, L. 2022.
\newblock Attention, filling in the gaps for generalization in routing problems.
\newblock In \emph{ECML PKDD}, 505--520. Springer.

\bibitem[{Bello et~al.(2016)Bello, Pham, Le, Norouzi, and Bengio}]{bello2016neural}
Bello, I.; Pham, H.; Le, Q.~V.; Norouzi, M.; and Bengio, S. 2016.
\newblock Neural combinatorial optimization with reinforcement learning.
\newblock \emph{arXiv preprint arXiv:1611.09940}.

\bibitem[{Bengio, Lodi, and Prouvost(2021)}]{bengio2021_ml4co_survey}
Bengio, Y.; Lodi, A.; and Prouvost, A. 2021.
\newblock Machine learning for combinatorial optimization: a methodological tour d’horizon.
\newblock \emph{European Journal of Operational Research}, 290(2): 405--421.

\bibitem[{Berto et~al.(2024{\natexlab{a}})Berto, Hua, Luttmann, Son, Park, Ahn, Kwon, Xie, and Park}]{berto2024parco}
Berto, F.; Hua, C.; Luttmann, L.; Son, J.; Park, J.; Ahn, K.; Kwon, C.; Xie, L.; and Park, J. 2024{\natexlab{a}}.
\newblock {PARCO: Learning Parallel Autoregressive Policies for Efficient Multi-Agent Combinatorial Optimization}.
\newblock \emph{arXiv preprint arXiv:2409.03811}.

\bibitem[{Berto et~al.(2025)Berto, Hua, Park, Luttmann, Ma, Bu, Wang, Ye, Kim, Choi, Zepeda, Hottung, Zhou, Bi, Hu, Liu, Kim, Son, Kim, Angioni, Kool, Cao, Zhang, Shin, Wu, Ahn, Song, Kwon, Xie, and Park}]{berto2023rl4co}
Berto, F.; Hua, C.; Park, J.; Luttmann, L.; Ma, Y.; Bu, F.; Wang, J.; Ye, H.; Kim, M.; Choi, S.; Zepeda, N.~G.; Hottung, A.; Zhou, J.; Bi, J.; Hu, Y.; Liu, F.; Kim, H.; Son, J.; Kim, H.; Angioni, D.; Kool, W.; Cao, Z.; Zhang, J.; Shin, K.; Wu, C.; Ahn, S.; Song, G.; Kwon, C.; Xie, L.; and Park, J. 2025.
\newblock {RL4CO: an Extensive Reinforcement Learning for Combinatorial Optimization Benchmark}.
\newblock In \emph{Proceedings of the 31st ACM SIGKDD Conference on Knowledge Discovery and Data Mining}.

\bibitem[{Berto et~al.(2024{\natexlab{b}})Berto, Hua, Zepeda, Hottung, Wouda, Lan, Tierney, and Park}]{berto2024routefinder}
Berto, F.; Hua, C.; Zepeda, N.~G.; Hottung, A.; Wouda, N.; Lan, L.; Tierney, K.; and Park, J. 2024{\natexlab{b}}.
\newblock {RouteFinder}: Towards Foundation Models for Vehicle Routing Problems.
\newblock In \emph{ICML 2024 FM-Wild Workshop}.

\bibitem[{Bi et~al.(2024)Bi, Ma, Zhou, Song, Cao, Wu, and Zhang}]{bi2024learning}
Bi, J.; Ma, Y.; Zhou, J.; Song, W.; Cao, Z.; Wu, Y.; and Zhang, J. 2024.
\newblock Learning to Handle Complex Constraints for Vehicle Routing Problems.
\newblock \emph{arXiv preprint arXiv:2410.21066}.

\bibitem[{Bogyrbayeva et~al.(2023)Bogyrbayeva, Yoon, Ko, Lim, Yun, and Kwon}]{bogyrbayeva2023deep}
Bogyrbayeva, A.; Yoon, T.; Ko, H.; Lim, S.; Yun, H.; and Kwon, C. 2023.
\newblock A deep reinforcement learning approach for solving the traveling salesman problem with drone.
\newblock \emph{Transportation Research Part C: Emerging Technologies}, 148: 103981.

\bibitem[{Braekers, Ramaekers, and Van~Nieuwenhuyse(2016)}]{braekers2016vehicle}
Braekers, K.; Ramaekers, K.; and Van~Nieuwenhuyse, I. 2016.
\newblock The vehicle routing problem: State of the art classification and review.
\newblock \emph{Computers \& industrial engineering}, 99: 300--313.

\bibitem[{Choo et~al.(2022)Choo, Kwon, Kim, Jae, Hottung, Tierney, and Gwon}]{choo2022simulation}
Choo, J.; Kwon, Y.-D.; Kim, J.; Jae, J.; Hottung, A.; Tierney, K.; and Gwon, Y. 2022.
\newblock Simulation-guided beam search for neural combinatorial optimization.
\newblock \emph{NeurIPS}, 35: 8760--8772.

\bibitem[{Cordeau and Laporte(2001)}]{Cordeau2001}
Cordeau, J.-F.; and Laporte, G. 2001.
\newblock A tabu search heuristic for the site dependent vehicle routing problem with time windows.
\newblock \emph{INFOR}, 39(4): 292--298.

\bibitem[{Drakulic, Michel, and Andreoli(2025)}]{drakulic2024goal}
Drakulic, D.; Michel, S.; and Andreoli, J.-M. 2025.
\newblock GOAL: A Generalist Combinatorial Optimization Agent Learning.
\newblock In \emph{ICLR}.

\bibitem[{Drakulic et~al.(2024)Drakulic, Michel, Mai, Sors, and Andreoli}]{drakulic2024bq}
Drakulic, D.; Michel, S.; Mai, F.; Sors, A.; and Andreoli, J.-M. 2024.
\newblock Bq-nco: Bisimulation quotienting for efficient neural combinatorial optimization.
\newblock \emph{NeurIPS}, 36.

\bibitem[{Duan et~al.(2020)Duan, Zhan, Hu, Gong, Wei, Zhang, and Xu}]{duanEfficientlySolvingPractical2020}
Duan, L.; Zhan, Y.; Hu, H.; Gong, Y.; Wei, J.; Zhang, X.; and Xu, Y. 2020.
\newblock Efficiently {{Solving}} the {{Practical Vehicle Routing Problem}}: {{A Novel Joint Learning Approach}}.
\newblock In \emph{KDD}. ACM.

\bibitem[{Goh et~al.(2025)Goh, Ma, Zhou, Cao, Dupty, and Lee}]{gohshield}
Goh, Y.~L.; Ma, Y.; Zhou, J.; Cao, Z.; Dupty, M.~H.; and Lee, W.~S. 2025.
\newblock SHIELD: Multi-task Multi-distribution Vehicle Routing Solver with Sparsity \& Hierarchy in Efficiently Layered Decoder.
\newblock In \emph{ICML}.

\bibitem[{Golden et~al.(1984)Golden, Assad, Levy, and Gheysens}]{golden1984fleet}
Golden, B.; Assad, A.; Levy, L.; and Gheysens, F. 1984.
\newblock The fleet size and mix vehicle routing problem.
\newblock \emph{Computers \& Operations Research}, 11(1): 49--66.

\bibitem[{Grinsztajn et~al.(2023)Grinsztajn, Furelos-Blanco, Surana, Bonnet, and Barrett}]{grinsztajn2023winner}
Grinsztajn, N.; Furelos-Blanco, D.; Surana, S.; Bonnet, C.; and Barrett, T. 2023.
\newblock Winner takes it all: Training performant RL populations for combinatorial optimization.
\newblock \emph{NeurIPS}, 36: 48485--48509.

\bibitem[{Grinsztajn et~al.(2024)Grinsztajn, Furelos-Blanco, Surana, Bonnet, and Barrett}]{grinsztajn2024winner}
Grinsztajn, N.; Furelos-Blanco, D.; Surana, S.; Bonnet, C.; and Barrett, T. 2024.
\newblock Winner Takes It All: Training Performant RL Populations for Combinatorial Optimization.
\newblock \emph{NeurIPS}, 36.

\bibitem[{Hottung, Kwon, and Tierney(2022)}]{hottung2021efficient}
Hottung, A.; Kwon, Y.-D.; and Tierney, K. 2022.
\newblock Efficient active search for combinatorial optimization problems.
\newblock In \emph{ICLR}.

\bibitem[{Hottung, Mahajan, and Tierney(2025)}]{hottung2024polynet}
Hottung, A.; Mahajan, M.; and Tierney, K. 2025.
\newblock {PolyNet}: Learning Diverse Solution Strategies for Neural Combinatorial Optimization.
\newblock \emph{ICLR}.

\bibitem[{Hottung and Tierney(2020)}]{hottung2020neural}
Hottung, A.; and Tierney, K. 2020.
\newblock Neural large neighborhood search for the capacitated vehicle routing problem.
\newblock In \emph{ECAI 2020}. IOS Press.

\bibitem[{Hottung, Wong-Chung, and Tierney(2025)}]{hottung2025neural}
Hottung, A.; Wong-Chung, P.; and Tierney, K. 2025.
\newblock Neural Deconstruction Search for Vehicle Routing Problems.
\newblock \emph{TMLR}.

\bibitem[{Hua et~al.(2025)Hua, Berto, Son, Kang, Kwon, and Park}]{hua2025camp_vrp}
Hua, C.; Berto, F.; Son, J.; Kang, S.; Kwon, C.; and Park, J. 2025.
\newblock {CAMP: Collaborative Attention Model with Profiles for Vehicle Routing Problems}.
\newblock In \emph{AAMAS}.

\bibitem[{Huang et~al.(2025)Huang, Zhou, Cao, and Xu}]{huang2025rethinking}
Huang, Z.; Zhou, J.; Cao, Z.; and Xu, Y. 2025.
\newblock Rethinking Light Decoder-based Solvers for Vehicle Routing Problems.
\newblock \emph{arXiv preprint arXiv:2503.00753}.

\bibitem[{Johnson and McGeoch(1997)}]{johnson1997traveling}
Johnson, D.~S.; and McGeoch, L.~A. 1997.
\newblock The traveling salesman problem: a case study.
\newblock \emph{Local search in combinatorial optimization}.

\bibitem[{Kim et~al.(2025)Kim, Choi, Son, Park, and Kwon}]{kim2025neural}
Kim, H.; Choi, S.; Son, J.; Park, J.; and Kwon, C. 2025.
\newblock Neural Genetic Search in Discrete Spaces.
\newblock In \emph{ICML}.

\bibitem[{Kim et~al.(2024)Kim, Choi, Son, Kim, Park, and Bengio}]{kim2024gfacs}
Kim, M.; Choi, S.; Son, J.; Kim, H.; Park, J.; and Bengio, Y. 2024.
\newblock Ant Colony Sampling with GFlowNets for Combinatorial Optimization.
\newblock \emph{arXiv preprint arXiv:2403.07041}.

\bibitem[{Kim, Park, and Park(2022{\natexlab{a}})}]{kim2022neuro}
Kim, M.; Park, J.; and Park, J. 2022{\natexlab{a}}.
\newblock Neuro CROSS exchange: Learning to CROSS exchange to solve realistic vehicle routing problems.
\newblock \emph{arXiv preprint arXiv:2206.02771}.

\bibitem[{Kim, Park, and Park(2022{\natexlab{b}})}]{kim2022sym}
Kim, M.; Park, J.; and Park, J. 2022{\natexlab{b}}.
\newblock Sym-nco: Leveraging symmetricity for neural combinatorial optimization.
\newblock \emph{NeurIPS}, 35: 1936--1949.

\bibitem[{Kingma and Ba(2014)}]{kingma2014adam}
Kingma, D.~P.; and Ba, J. 2014.
\newblock Adam: A method for stochastic optimization.
\newblock \emph{arXiv preprint arXiv:1412.6980}.

\bibitem[{Kool et~al.(2022)Kool, van Hoof, Gromicho, and Welling}]{kool2022deep}
Kool, W.; van Hoof, H.; Gromicho, J.; and Welling, M. 2022.
\newblock Deep policy dynamic programming for vehicle routing problems.
\newblock In \emph{CPAIOR}, 190--213. Springer.

\bibitem[{Kool, Van~Hoof, and Welling(2018)}]{kool2018attention}
Kool, W.; Van~Hoof, H.; and Welling, M. 2018.
\newblock Attention, learn to solve routing problems!
\newblock \emph{arXiv preprint arXiv:1803.08475}.

\bibitem[{Kwon et~al.(2020)Kwon, Choo, Kim, Yoon, Gwon, and Min}]{kwon2020pomo}
Kwon, Y.-D.; Choo, J.; Kim, B.; Yoon, I.; Gwon, Y.; and Min, S. 2020.
\newblock Pomo: Policy optimization with multiple optima for reinforcement learning.
\newblock \emph{NeurIPS}, 33: 21188--21198.

\bibitem[{Li et~al.(2024{\natexlab{a}})Li, Liu, Zheng, Zhang, and Wang}]{li2024cada}
Li, H.; Liu, F.; Zheng, Z.; Zhang, Y.; and Wang, Z. 2024{\natexlab{a}}.
\newblock CaDA: Cross-Problem Routing Solver with Constraint-Aware Dual-Attention.
\newblock \emph{arXiv preprint arXiv:2412.00346}.

\bibitem[{Li et~al.(2022)Li, Ma, Gao, Cao, Lim, Song, and Zhang}]{li2022deep}
Li, J.; Ma, Y.; Gao, R.; Cao, Z.; Lim, A.; Song, W.; and Zhang, J. 2022.
\newblock Deep reinforcement learning for solving the heterogeneous capacitated vehicle routing problem.
\newblock \emph{IEEE Transactions on Cybernetics}.

\bibitem[{Li, Yan, and Wu(2021)}]{li2021learning}
Li, S.; Yan, Z.; and Wu, C. 2021.
\newblock Learning to delegate for large-scale vehicle routing.
\newblock \emph{NeurIPS}, 34: 26198--26211.

\bibitem[{Li et~al.(2024{\natexlab{b}})Li, Guo, Wang, and Yan}]{li2024distribution}
Li, Y.; Guo, J.; Wang, R.; and Yan, J. 2024{\natexlab{b}}.
\newblock From distribution learning in training to gradient search in testing for combinatorial optimization.
\newblock \emph{NeurIPS}, 36.

\bibitem[{Li et~al.(2025)Li, Ma, Pan, Wang, Geng, Yang, and Yan}]{li2025unify}
Li, Y.; Ma, J.; Pan, W.; Wang, R.; Geng, H.; Yang, N.; and Yan, J. 2025.
\newblock Unify ml4tsp: Drawing methodological principles for tsp and beyond from streamlined design space of learning and search.
\newblock In \emph{The Thirteenth International Conference on Learning Representations}.

\bibitem[{Li et~al.(2023)Li, Zhou, Yuan, and Ngo}]{li2023experience}
Li, Y.; Zhou, C.; Yuan, P.; and Ngo, T. T.~A. 2023.
\newblock Experience-based territory planning and driver assignment with predicted demand and driver present condition.
\newblock \emph{Transportation research part E: logistics and transportation review}, 171: 103036.

\bibitem[{Liu et~al.(2024{\natexlab{a}})Liu, Lin, Wang, Zhang, Xialiang, and Yuan}]{liu2024multi}
Liu, F.; Lin, X.; Wang, Z.; Zhang, Q.; Xialiang, T.; and Yuan, M. 2024{\natexlab{a}}.
\newblock Multi-task learning for routing problem with cross-problem zero-shot generalization.
\newblock In \emph{KDD}.

\bibitem[{Liu et~al.(2024{\natexlab{b}})Liu, Xialiang, Yuan, Lin, Luo, Wang, Lu, and Zhang}]{liu2024evolution}
Liu, F.; Xialiang, T.; Yuan, M.; Lin, X.; Luo, F.; Wang, Z.; Lu, Z.; and Zhang, Q. 2024{\natexlab{b}}.
\newblock Evolution of Heuristics: Towards Efficient Automatic Algorithm Design Using Large Language Model.
\newblock In \emph{ICML}.

\bibitem[{Liu et~al.(2024{\natexlab{c}})Liu, Liu, Niu, Long, Zhang, and Xu}]{liu20242d}
Liu, Q.; Liu, C.; Niu, S.; Long, C.; Zhang, J.; and Xu, M. 2024{\natexlab{c}}.
\newblock 2D-Ptr: 2D Array Pointer Network for Solving the Heterogeneous Capacitated Vehicle Routing Problem.
\newblock In \emph{AAMAS}, 1238--1246.

\bibitem[{Lozano, Molina, and Herrera(2011)}]{lozano2011editorial}
Lozano, M.; Molina, D.; and Herrera, F. 2011.
\newblock Editorial scalability of evolutionary algorithms and other metaheuristics for large-scale continuous optimization problems.
\newblock \emph{Soft computing}, 15: 2085--2087.

\bibitem[{Ma, Cao, and Chee(2024)}]{ma2024learning}
Ma, Y.; Cao, Z.; and Chee, Y.~M. 2024.
\newblock Learning to search feasible and infeasible regions of routing problems with flexible neural k-opt.
\newblock \emph{NeurIPS}, 36.

\bibitem[{Mazyavkina et~al.(2021)Mazyavkina, Sviridov, Ivanov, and Burnaev}]{mazyavkina2021reinforcement}
Mazyavkina, N.; Sviridov, S.; Ivanov, S.; and Burnaev, E. 2021.
\newblock Reinforcement learning for combinatorial optimization: A survey.
\newblock \emph{Computers \& Operations Research}, 134: 105400.

\bibitem[{Ouyang et~al.(2025)Ouyang, Li, Ma, and Wu}]{ouyang2025learning}
Ouyang, W.; Li, S.; Ma, Y.; and Wu, C. 2025.
\newblock Learning to Segment for Capacitated Vehicle Routing Problems.

\bibitem[{Papadimitriou and Steiglitz(1998)}]{papadimitriou1998combinatorial}
Papadimitriou, C.~H.; and Steiglitz, K. 1998.
\newblock \emph{Combinatorial optimization: algorithms and complexity}.
\newblock Courier Corporation.

\bibitem[{Perron and Furnon(2023)}]{perron2023ortools}
Perron, L.; and Furnon, V. 2023.
\newblock {{OR-Tools}}.
\newblock Google.

\bibitem[{Pham, Doan, and Huynh(2025)}]{pham2025hsevo}
Pham, V. T.~D.; Doan, L.; and Huynh, T. T.~B. 2025.
\newblock {HSEvo: Elevating Automatic Heuristic Design with Diversity-Driven Harmony Search and Genetic Algorithm Using LLMs}.
\newblock In \emph{AAAI}. Association for the Advancement of Artificial Intelligence (AAAI).

\bibitem[{Pirnay and Grimm(2024)}]{pirnay2024take}
Pirnay, J.; and Grimm, D.~G. 2024.
\newblock Take a step and reconsider: Sequence decoding for self-improved neural combinatorial optimization.
\newblock In \emph{ECAI}.

\bibitem[{Son et~al.(2024)Son, Kim, Choi, Kim, and Park}]{son2024equity}
Son, J.; Kim, M.; Choi, S.; Kim, H.; and Park, J. 2024.
\newblock Equity-Transformer: Solving NP-Hard Min-Max Routing Problems as Sequential Generation with Equity Context.
\newblock In \emph{AAAI}.

\bibitem[{Son et~al.(2025)Son, Zhao, Berto, Hua, Kwon, and Park}]{son2025neural}
Son, J.; Zhao, Z.; Berto, F.; Hua, C.; Kwon, C.; and Park, J. 2025.
\newblock Neural Combinatorial Optimization for Real-World Routing.
\newblock \emph{arXiv preprint arXiv:2503.16159}.

\bibitem[{Sun et~al.(2019)Sun, Li, Wang, He, Lin, and Deng}]{sun2019fast}
Sun, Z.; Li, Z.; Wang, H.; He, D.; Lin, Z.; and Deng, Z. 2019.
\newblock Fast structured decoding for sequence models.
\newblock \emph{NeurIPS}, 32.

\bibitem[{Sun and Yang(2024)}]{sun2024difusco}
Sun, Z.; and Yang, Y. 2024.
\newblock Difusco: Graph-based diffusion solvers for combinatorial optimization.
\newblock \emph{NeurIPS}, 36.

\bibitem[{{Team Locus}(2020)}]{locus2020zonebased}
{Team Locus}. 2020.
\newblock Zone-Based Routing is the Need of the Hour.
\newblock \emph{Locus Blog}.
\newblock Access: 2024-10-17.

\bibitem[{Tran et~al.(2025)Tran, Nguyen-Tri, Binh, and Thanh-Tung}]{tranlarge}
Tran, C.~D.; Nguyen-Tri, Q.; Binh, H. T.~T.; and Thanh-Tung, H. 2025.
\newblock Large Language Models powered Neural Solvers for Generalized Vehicle Routing Problems.
\newblock In \emph{Towards Agentic AI for Science: Hypothesis Generation, Comprehension, Quantification, and Validation}.

\bibitem[{Vidal(2022)}]{vidal2022hybrid}
Vidal, T. 2022.
\newblock Hybrid genetic search for the CVRP: Open-source implementation and SWAP* neighborhood.
\newblock \emph{Computers \& Operations Research}, 140: 105643.

\bibitem[{Vinyals, Fortunato, and Jaitly(2015)}]{vinyals2015pointer}
Vinyals, O.; Fortunato, M.; and Jaitly, N. 2015.
\newblock Pointer networks.
\newblock \emph{NeurIPS}, 28.

\bibitem[{Wang et~al.(2024)Wang, Jia, Chen, and Mei}]{wang2024distance}
Wang, Y.; Jia, Y.-H.; Chen, W.-N.; and Mei, Y. 2024.
\newblock Distance-aware Attention Reshaping: Enhance Generalization of Neural Solver for Large-scale Vehicle Routing Problems.
\newblock \emph{arXiv preprint arXiv:2401.06979}.

\bibitem[{Wouda, Lan, and Kool(2024)}]{wouda2024pyvrp}
Wouda, N.~A.; Lan, L.; and Kool, W. 2024.
\newblock {P}y{VRP}: A high-performance {VRP} solver package.
\newblock \emph{INFORMS Journal on Computing}.

\bibitem[{Ye et~al.(2024)Ye, Wang, Cao, Berto, Hua, Kim, Park, and Song}]{ye2024reevo}
Ye, H.; Wang, J.; Cao, Z.; Berto, F.; Hua, C.; Kim, H.; Park, J.; and Song, G. 2024.
\newblock ReEvo: Large Language Models as Hyper-Heuristics with Reflective Evolution.
\newblock In \emph{NeurIPS}.

\bibitem[{Ye et~al.(2023)Ye, Wang, Cao, Liang, and Li}]{ye2023deepaco}
Ye, H.; Wang, J.; Cao, Z.; Liang, H.; and Li, Y. 2023.
\newblock DeepACO: Neural-enhanced Ant Systems for Combinatorial Optimization.
\newblock In \emph{NeurIPS}.

\bibitem[{Zhang et~al.(2025)Zhang, Yang, Cao, and Chi}]{zhang2025adversarial}
Zhang, N.; Yang, J.; Cao, Z.; and Chi, X. 2025.
\newblock Adversarial Generative Flow Network for Solving Vehicle Routing Problems.
\newblock \emph{arXiv preprint arXiv:2503.01931}.

\bibitem[{Zhao et~al.(2025)Zhao, Hua, Berto, Lee, Ma, Li, and Park}]{zhao2025trajevo}
Zhao, Z.; Hua, C.; Berto, F.; Lee, K.; Ma, Z.; Li, J.; and Park, J. 2025.
\newblock {TrajEvo}: Designing Trajectory Prediction Heuristics via LLM-driven Evolution.
\newblock \emph{arXiv preprint arXiv:2505.04480}.

\bibitem[{Zheng et~al.(2024)Zheng, Yao, Wang, Xialiang, Yuan, and Tang}]{zheng2024dpn}
Zheng, Z.; Yao, S.; Wang, Z.; Xialiang, T.; Yuan, M.; and Tang, K. 2024.
\newblock {DPN}: Decoupling Partition and Navigation for Neural Solvers of Min-max Vehicle Routing Problems.
\newblock In \emph{ICML}.

\bibitem[{Zhong, Hall, and Dessouky(2007)}]{zhong2007territory}
Zhong, H.; Hall, R.~W.; and Dessouky, M. 2007.
\newblock Territory planning and vehicle dispatching with driver learning.
\newblock \emph{Transportation Science}.

\bibitem[{Zhou et~al.(2024)Zhou, Cao, Wu, Song, Ma, Zhang, and Xu}]{zhou2024mvmoe}
Zhou, J.; Cao, Z.; Wu, Y.; Song, W.; Ma, Y.; Zhang, J.; and Xu, C. 2024.
\newblock MVMoE: Multi-Task Vehicle Routing Solver with Mixture-of-Experts.
\newblock In \emph{ICML}.

\bibitem[{Zong et~al.(2022)Zong, Zheng, Li, and Jin}]{zong2022mapdp}
Zong, Z.; Zheng, M.; Li, Y.; and Jin, D. 2022.
\newblock Mapdp: Cooperative multi-agent reinforcement learning to solve pickup and delivery problems.
\newblock In \emph{AAAI}.

\end{thebibliography}

\clearpage

\onecolumn
\appendix

\section*{Supplementary Materials}

\subsection*{Detailed PVRP Definition}

We consider a directed graph $G=(V,E)$, where $V = \{0, \dots, N\}$ is the set of nodes, including $\{0\}$ as the depot and $\{1, \dots, N\}$ as clients, and vehicle set $K = \{1,\dots,M\}$ is the set of vehicles. Each client $i\in V$ has demand $d_i$, and each vehicle $k\in K$ has capacity $Q_k$ and speed $v_k$. Between each client $i$ and vehicle $k$, there is a profile score $p_{ik}\in\mathbb{R}\cup\{\pm\infty\}$. Intuitively, a higher profile score means that vehicle $k$ is encouraged to serve client $i$; symmetrically, this could also be understood as client $i$ preferring vehicle $k$ to serve them. Particularly, if the profile score is $\pm\infty$, it represents a hard constraint: $-\infty$ means that vehicle $k$ cannot serve client $i$, while $+\infty$ means that vehicle $k$ must serve client $i$. The edges $E$ connect pairs of nodes, and each edge between node $i$ and node $j$, $(i, j) \in E$, has a travel distance $c_{ij}$.

We introduce the decision variables $x^k_{ij}=1$ if vehicle $k$ travels from $i$ to $j$ (and $0$ otherwise) and $y^k_i=1$ if vehicle $k$ serves client $i$ (and $0$ otherwise). The objective is to maximize total profile reward minus travel time, balanced via a \textit{profile weight} $\alpha \in [0, 1]$:
\begin{equation}
\label{eq:objective-supplementary}
\max_{x,y}\;\sum_{k\in K}\sum_{i\in V}\sum_{j\in V}
\bigl(\alpha\,p_{ik}-\tfrac{c_{ij}}{v_k}\bigr)\,x^k_{ij}
\end{equation}

This optimization is subject to the following constraints:
\begin{align}
&\sum_{k\in K}y_i^k = 1
&&\forall\,i\in \{1,\dots,N\},
\tag{2a}\\
&\sum_{i\in \{1,\dots,N\}}\sum_{j\in V}d_i\,x^k_{ij} \le Q_k
&&\forall\,k\in K,
\tag{2b}\\
&\sum_{j\in V}x^k_{hj} = \sum_{i\in V}x^k_{ih}
&&\forall\,h\in V,\;k\in K,
\tag{2c}\\
&x^k_{ij}\in\{0,1\},\quad y_i^k\in\{0,1\}
&&\forall\,i,j\in V,\;k\in K.
\tag{2d}
\end{align}

\paragraph{Client Service Constraint (2a)} This constraint ensures that each client is served exactly once across all vehicles. The summation over all vehicles $k \in K$ for each client $i$ must equal 1, guaranteeing that no client is left unserved and no client receives redundant service. This is a fundamental requirement in vehicle routing problems to maintain service completeness.

\paragraph{Vehicle Capacity Constraint (2b)} This constraint enforces that the total demand served by each vehicle $k$ does not exceed its capacity $Q_k$. The constraint considers the demand $d_i$ of each client $i$ that vehicle $k$ visits (indicated by the decision variable $x^k_{ij}$ where the vehicle travels from client $i$ to any destination $j$). This ensures operational feasibility by preventing vehicles from being overloaded beyond their physical or regulatory limits.

\paragraph{Flow Conservation Constraint (2c)} This constraint maintains route continuity and prevents the formation of subtours. For each node $h$ (including the depot and all clients) and each vehicle $k$, the number of edges entering node $h$ must equal the number of edges leaving node $h$. This ensures that if a vehicle arrives at a location, it must also depart from that location, creating valid continuous routes. Additionally, this constraint implicitly ensures that each vehicle's route forms a single connected tour starting and ending at the depot.

\paragraph{Binary Constraint (2d)} This constraint enforces the binary nature of all decision variables. The routing variables $x^k_{ij}$ can only take values 0 or 1, indicating whether vehicle $k$ travels from node $i$ to node $j$. Similarly, the service variables $y^k_i$ are binary, indicating whether vehicle $k$ serves client $i$. This integrality constraint is essential for maintaining the discrete optimization nature of the problem and ensuring that the solution represents actual routing decisions.

\paragraph{Problem variants} By setting all profile scores $p_{ik}=0$, we recover the classical Capacitated Vehicle Routing Problem (CVRP) where the objective is to minimize the total travel time. The introduction of non-zero profile scores allows the PVRP to capture various real-world scenarios including client-vehicle preferences, vehicle specialization requirements, and zone-based service restrictions through the use of infinite profile scores as hard constraints.

\subsection{Detailed MDP Formulation}

The PVRP can be naturally framed as a Markov Decision Process (MDP), which enables the application of reinforcement learning techniques for scalable and adaptive solution generation. We define the MDP formulation as a 5-tuple $(\mathcal{S}, \mathcal{A}, \mathcal{T}, \mathcal{R}, \pi)$ with the following components:

\paragraph{State Space ($\mathcal{S}$)} A state $s^t \in \mathcal{S}$ at time step $t$ captures the partial route constructed up to that point. Specifically, the state includes the current location of all vehicles $s^t_k \in \mathbb{R}^2$ for each vehicle $k \in K$, the remaining capacity $Q_k^t$ of each vehicle, the set of visited and unvisited clients, the client-vehicle profile matrix $P=\{p_{ik}, i\in V, k\in K\}\in\mathbb{R}^{N\times M}$, and the accumulated travel cost. This comprehensive state representation provides a complete view of the current routing status and enables the policy to make informed decisions about subsequent actions. The state space is finite but exponentially large, as it encompasses all possible combinations of vehicle positions, capacity states, and client visit patterns.

\paragraph{Action Space ($\mathcal{A}$)} An action $a\in\mathcal{A}$ consists of selecting the client to visit next or returning to the depot. At each decision step, the action space encompasses all unvisited clients $i \in \{1,\dots,N\}$ that satisfy capacity constraints and zone restrictions for the current vehicle, as well as the option to return to the depot to initialize a new route segment when necessary. Formally, for a vehicle $k$ at time $t$, the feasible action set is defined as:
\begin{equation}
\mathcal{A}^t_k = \{i \in \{1,\dots,N\} : i \text{ unvisited}, d_i \leq Q_k^t, p_{ik} > -\infty\} \cup \{0\}
\end{equation}
where the depot node $\{0\}$ is always available as an action to terminate the current route segment. The action space is dynamic and shrinks as clients are served and vehicle capacities are consumed.

\paragraph{Transition Dynamics ($\mathcal{T}$)} The system evolves according to a deterministic transition function $s_{t+1} = \mathcal{T}(s_t, a_t)$, which updates vehicle locations, remaining capacities, and the set of visited clients based on the selected action. When a vehicle visits a client $i$, the client's demand $d_i$ is subtracted from the vehicle's remaining capacity, the client is marked as served and removed from the unvisited set, and the vehicle's location is updated to the client's coordinates. If a vehicle returns to the depot, its capacity is reset to the initial value $Q_k$, enabling it to start a new route segment. The transition function ensures that all problem constraints are maintained throughout the solution construction process.

\paragraph{Reward Function ($\mathcal{R}$)} To align with the bi-objective nature of PVRP, we define the reward for each action as:
\begin{equation}
r_t = \alpha p_{jk} - \frac{c_{ij}}{v_k}
\end{equation}
as per the objective function in \cref{eq:objective}, where $p_{jk}$ represents the client-vehicle preference score, $c_{ij}/v_k$ accounts for the travel cost (time), and $\alpha$ controls the trade-off between preference satisfaction and transportation efficiency. In practice, we employ a sparse reward strategy, calculating the cumulative reward efficiently at the end of each episode when solution construction completes. This approach is equivalent to using the unified objective defined in the problem formulation and helps stabilize the learning process by avoiding frequent intermediate reward signals that might mislead the policy during training.

\paragraph{Policy ($\pi$)} A policy $\pi(a_t | s_t)$ specifies the probability distribution over possible actions given the current state. Our objective is to learn an optimal policy $\pi^*$ that maximizes the expected cumulative reward:
\begin{equation}
\pi^* = \arg\max_{\pi} \mathbb{E}\left[ \sum_{t=0}^{T} \gamma^t r_t \right]
\end{equation}
where $\gamma \in [0,1]$ is a discount factor which we set to $\gamma=1$ (effectively no discounting) due to the sparse nature of our reward signal. The policy is parameterized by a neural network that learns to map states to action probabilities, enabling the agent to make sequential routing decisions that optimize the overall objective while satisfying all problem constraints.

\paragraph{Episode Termination} An episode terminates when all clients have been served exactly once and all vehicles have returned to the depot. The final state represents a complete feasible solution to the PVRP instance, and the cumulative reward corresponds to the objective value of this solution. The MDP formulation naturally handles the sequential decision-making nature of vehicle routing while incorporating the preference-based considerations that distinguish PVRP from classical VRP variants.

\subsection{Additional Experimental Details}

\paragraph{Code Implementation and Hardware} Our code is implemented in PyTorch using the RL4CO framework \citep{berto2023rl4co}, which provides a comprehensive library for reinforcement learning-based combinatorial optimization. We are committed to releasing the complete source code and trained models upon acceptance to foster reproducible academic research. All training and testing experiments are conducted on an AMD Ryzen Threadripper 3960X (24-core) CPU with an NVIDIA RTX 4090 GPU (24GB VRAM). The computational infrastructure ensures consistent and reliable experimental conditions across all evaluated methods.

\paragraph{Data Generation} For the basic features of VRP instances, including client and depot coordinates, demands, vehicle capacity, and speed, we follow the widely used settings established by \citet{kool2018attention}. Specifically, we generate synthetic PVRP instances by sampling client and depot coordinates from $\sim\mathrm{Uniform}(0,1)$ and drawing client demands from $\sim\mathrm{UniformInteger}(1,9)$. We use $M=7$ vehicles and $N=100$ clients in the main experiments, where each vehicle has capacity $Q_k=40$ and speed $v_k=1$, ensuring that Euclidean distance equals travel time for computational simplicity.

The key distinguishing feature of PVRP instances lies in the profile score generation. Profile scores are drawn independently for all client-vehicle pairs as $p_{ik} \sim\mathrm{Uniform}(0,1)$. To introduce hard constraints that reflect real-world scenarios, we sample two probabilities from $\sim\mathrm{Uniform}(0,0.1)$ for each instance to randomly set portions of the profile matrix to $\pm\infty$. The first probability determines the fraction of client-vehicle pairs assigned $p_{ik}=-\infty$ (forbidden assignments), while the second probability determines those assigned $p_{ik}=+\infty$ (mandatory assignments). This approach creates diverse constraint patterns without losing generalization capability.

For zone-constrained variants, we partition the service area into $S = M$ angular sectors centered at the depot. We sample the constraint rate and for each client $i$, randomly mask $\lfloor rS\rfloor$ sectors. Vehicles whose home sector is masked for client $i$ are assigned $p_{ik}=-\infty$, effectively forbidding those assignments and creating realistic geographical service restrictions.

\paragraph{Training Configuration} All models are trained under identical settings to ensure fair comparison. We optimize using the Adam optimizer \citep{kingma2014adam} with an initial learning rate of $10^{-4}$, which decays by a factor of 0.1 at epochs 80 and 95. Training runs for 100 epochs with $10^5$ samples per epoch, using a batch size of 32. For baseline estimation, we employ 8 augmented rollouts per instance via Sym-NCO \citep{kim2022sym}, which provides stable gradient estimates for policy gradient methods.

Architecturally, each model employs $d_h=128$ hidden-dimensional embeddings, 8 attention heads, and 512-dimensional feedforward layers across 3 encoder layers. These architectural choices balance model expressiveness with computational efficiency. During training, we scale the number of clients from 60 to 100 and the number of vehicles from 3 to 7, providing diverse problem sizes for robust learning.

For baseline models, following the established setting from CAMP, we train separate models for each fixed $\alpha$ value taken from $\alpha \in \{0.00, 0.025, \ldots, 0.20\}$, where the final reported performance represents the average of results across all $\alpha$ values. In contrast, as a unified model, our approach is trained with $\alpha$ randomly sampled from the interval $[0, 0.2]$ for each instance, also incorporating a sampling probability for mixed hard-constraints ranging from 0\% to 10\%. This unified training strategy enables our model to handle the full spectrum of preference-cost trade-offs without requiring separate model training.

\paragraph{Testing Protocol} We evaluate our approach across three comprehensive settings to assess both performance and generalization capability. First, for in-distribution evaluation, we test each model on 1,280 randomly generated instances for each problem size, following the same generation rules as training data. This provides a robust assessment of model performance under expected conditions.

Second, we conduct out-of-distribution generalization tests using 128 instances with larger vehicle numbers $M\in\{15,25,35\}$ for $N=500$ clients. This evaluation assesses the model's ability to scale beyond training distributions and handle larger, more complex problem instances.

Finally, we introduce PVRPLib, a variation of real-world data derived from CVRPLib\footnote{\url{http://vrp.atd-lab.inf.puc-rio.br/index.php/en/}}. PVRPLib retains the number of vehicles, capacity values, and coordinates from the original CVRPLib instances but incorporates profile scores generated according to our data generation protocol. This hybrid approach combines realistic geographical and logistical constraints with the preference-based considerations that define PVRP.

For each instance, we measure both greedy performance (denoted as $g.$), obtained by taking the $\arg\max$ over decoder log-probabilities, and sampling performance (denoted as $s.$), which involves sampling 1,280 solutions per instance and selecting the best. Final performance metrics are reported as averages over all profile distributions and $\alpha$ settings, providing comprehensive evaluation coverage.

\subsection{Hyperparameters}

The experimental framework employs several hyperparameters that govern the neural network training process, reinforcement learning dynamics, and evaluation protocols. Key parameters used in our experiments are detailed in \cref{tab:hyperparams}. These settings include optimization configurations, architectural choices, training dynamics, and evaluation parameters. Many of these values follow established practices in neural combinatorial optimization and were empirically validated for the PVRP domain through preliminary experiments.

\begin{table}[htbp]
    \centering
    \small 
    \begin{tabular}{@{} p{5.5cm} l @{}}
        \toprule
        \textbf{Hyperparameter} & \textbf{Value} \\
        \midrule
        
        \multicolumn{2}{@{}l}{\textbf{Optimization}} \\
        \quad Optimizer & Adam \\
        \quad Initial learning rate & $10^{-4}$ \\
        \quad Learning rate decay factor & 0.1 \\
        \quad Decay epochs & 80, 95 \\
        \quad Total training epochs & 100 \\
        \quad Batch size & 32 \\
        \addlinespace
        
        \multicolumn{2}{@{}l}{\textbf{Neural Architecture}} \\
        \quad Hidden dimension ($d_h$) & 128 \\
        \quad Number of attention heads & 8 \\
        \quad Feedforward dimension & 512 \\
        \quad Number of encoder layers & 3 \\
        \addlinespace
        
        \multicolumn{2}{@{}l}{\textbf{Training Dynamics}} \\
        \quad Samples per epoch & $10^5$ \\
        \quad Augmented rollouts (Sym-NCO) & 8 \\
        \quad Client range (training) & 60-100 \\
        \quad Vehicle range (training) & 3-7 \\
        \quad Profile weight range ($\alpha$) & 0.0-0.2 \\
        \quad Hard constraint probability & 0\%-10\% \\
        \addlinespace
        
        \multicolumn{2}{@{}l}{\textbf{Problem Generation}} \\
        \quad Coordinate distribution & Uniform(0,1) \\
        \quad Demand distribution & UniformInteger(1,9) \\
        \quad Vehicle capacity ($Q_k$) & 40 \\
        \quad Vehicle speed ($v_k$) & 1.0 \\
        \quad Profile score distribution & Uniform(0,1) \\
        \addlinespace
        
        \multicolumn{2}{@{}l}{\textbf{Evaluation}} \\
        \quad In-distribution test instances & 1,280 \\
        \quad Out-of-distribution test instances & 128 \\
        \quad Sampling solutions per instance & 1,280 \\
        \bottomrule
    \end{tabular}
    \caption{Main hyperparameters for the PVRP framework.}
    \label{tab:hyperparams}
\end{table}

\subsection{Licenses for used assets}
\label{app:licenses-for-used-assets}

\cref{tab:asset} lists the used assets and their licenses. Our code is licensed under the MIT License.

\begin{table}[htbp]
    \centering
    \small 
    \begin{tabular}{@{} p{6cm} p{6cm} @{}}
        \toprule
        \textbf{Asset} & \textbf{License \& Usage} \\
        \midrule
        
        \multicolumn{2}{@{}l}{\textbf{Neural Baselines}} \\
        \quad ET \citep{son2024equity} & MIT License (Baseline) \\
        \quad DPN \citep{zheng2024dpn} & MIT License (Baseline) \\
        \quad 2D-Ptr \citep{liu20242d} & MIT License (Baseline) \\
        \quad PARCO \citep{berto2024parco} & MIT License (Baseline) \\
        \quad CAMP \citep{hua2025camp_vrp} & MIT License (Baseline) \\
        \addlinespace
        
        \multicolumn{2}{@{}l}{\textbf{Classical Solvers}} \\
        \quad OR-Tools \citep{perron2023ortools} & Apache-2.0 (Classical Solver) \\
        \quad HGS-PyVRP \citep{wouda2024pyvrp} & MIT License (Classical Solver) \\
        \addlinespace
        
        \multicolumn{2}{@{}l}{\textbf{Framework \& Libraries}} \\
        \quad RL4CO \citep{berto2023rl4co} & MIT License (Framework) \\
        \addlinespace
        
        \multicolumn{2}{@{}l}{\textbf{Datasets}} \\
        \quad CVRPLib & Non-commercial use (Testing) \\
        \quad Synthetic PVRP instances & Self-generated (Training \& Testing) \\
        \bottomrule
    \end{tabular}
    \caption{Used assets and their licenses.}
    \label{tab:asset}
\end{table}

\subsection{Additional Experimental Results}

\paragraph{Additional Out-of-Distribution Results}
The comprehensive evaluation across different problem sizes demonstrates the consistent superiority and strong generalization capability of our unified approach. Tables 8 and 9 present detailed results for medium-scale ($N=200$) and large-scale ($N=500$) PVRP instances, extending the analysis beyond the $N=1000$ results shown in the main paper. These results validate that our model maintains significant performance advantages across the entire spectrum of problem sizes, with gaps to the classical HGS-PyVRP solver consistently outperforming all neural baselines. Notably, our approach achieves the best performance-efficiency trade-off, delivering superior solution quality while maintaining fast inference times across all tested scales.

\begin{table}[h!]
\centering
\small
\setlength{\tabcolsep}{1mm}
\begin{tabular}{l ccc c l}
\toprule
M & 6 & 10 & 14 & Time & Gap(\%) \\
\midrule
OR-Tools & 17.74 & 19.80 & 20.61 & 15m & 9.06 \\
HGS-PyVRP & 16.73 & 17.32 & 18.24 & 15m & 0.00 \\
\midrule
ET (\textit{g.}) & 32.14 & 32.16 & 32.40 & 0.54s & 84.94 (+72.91)\\
DPN (\textit{g.}) & 27.95 & 27.97 & 28.17 & 0.54s & 60.81 (+48.78)\\
2D-Ptr (\textit{g.}) & 24.62 & 24.37 & 25.04 & 0.35s & 41.58 (+29.55)\\
PARCO (\textit{g.}) & 21.91 & 21.69 & 22.28 & 0.40s & 26.00 (+13.97)\\
CAMP (\textit{g.}) & 21.53 & 21.31 & 21.90 & 0.43s & 23.81 (+11.78) \\
\our{} (\textit{g.}) & \textbf{20.47} & \textbf{20.27} & \textbf{20.82} & \textbf{0.35s} & \textbf{17.75} \\
\midrule
ET (\textit{s.}) & 31.49 & 31.52 & 31.75 & 0.60s & 81.23 (+69.20)\\
DPN (\textit{s.}) & 24.79 & 24.80 & 24.76 & 0.60s & 42.21 (+30.18)\\
2D-Ptr (\textit{s.}) & 21.85 & 21.61 & 22.01 & 0.38s & 25.21 (+13.18)\\
PARCO (\textit{s.}) & 21.48 & 21.26 & 21.84 & 0.44s & 23.49 (+11.46)\\
CAMP (\textit{s.}) & 21.16 & 20.95 & 21.52 & 0.46s & 21.68 (+9.65)\\
\our{} (\textit{s.}) & \textbf{19.48} & \textbf{19.29} & \textbf{19.81} & \textbf{0.39s} & \textbf{12.03} \\
\bottomrule
\end{tabular}
\label{tab:mediumscale}
\caption{Benchmarks and results for medium-scale PVRP instances ($N=200$). We report the solution cost ($\downarrow$) and average gap ($\downarrow$) to the HGS-PyVRP solver. Average inference time is shown in the Time column.}
\vspace{-3mm}
\end{table}

\begin{table}[h!]
\centering
\small
\setlength{\tabcolsep}{1mm}
\begin{tabular}{l ccc c l}
\toprule
M & 15 & 25 & 35 & Time & Gap(\%) \\
\midrule
OR-Tools & 54.15 & 55.16 & 58.34 & 15m & 51.76 \\
HGS-PyVRP & 35.67 & 37.05 & 38.60 & 15m & 0.00 \\
\midrule
ET (\textit{g.}) & 73.91 & 73.71 & 74.31 & 1.13s & 99.36 (+86.39)\\
DPN (\textit{g.}) & 64.27 & 64.09 & 64.62 & 1.13s & 73.36 (+60.39)\\
2D-Ptr (\textit{g.}) & 53.01 & 52.86 & 53.09 & 0.60s & 42.80 (+29.83)\\
PARCO (\textit{g.}) & 47.18 & 47.04 & 47.25 & 0.66s & 27.08 (+14.11)\\
CAMP (\textit{g.}) & 46.35 & 46.22 & 46.43 & 0.72s & 24.87 (+11.90) \\
\our{} (\textit{g.}) & \textbf{44.09} & \textbf{43.96} & \textbf{44.15} & \textbf{0.59s} & \textbf{18.75} \\
\midrule
ET (\textit{s.}) & 72.43 & 72.23 & 72.83 & 1.27s & 95.37 (+82.40)\\
DPN (\textit{s.}) & 59.38 & 59.83 & 60.78 & 1.25s & 61.69 (+48.72)\\
2D-Ptr (\textit{s.}) & 48.99 & 49.34 & 49.93 & 0.67s & 33.18 (+20.21)\\
PARCO (\textit{s.}) & 46.24 & 46.10 & 46.31 & 0.73s & 24.55 (+11.58)\\
CAMP (\textit{s.}) & 45.55 & 45.42 & 45.62 & 0.78s & 22.71 (+9.74)\\
\our{} (\textit{s.}) & \textbf{41.94} & \textbf{41.82} & \textbf{42.00} & \textbf{0.66s} & \textbf{12.97} \\
\bottomrule
\end{tabular}
\label{tab:largescale500}
\caption{Benchmarks and results for large-scale PVRP instances ($N=500$). We report the solution cost ($\downarrow$) and average gap ($\downarrow$) to the HGS-PyVRP solver. Average inference time is shown in the Time column.}
\vspace{-3mm}
\end{table}

\end{document}